\pdfoutput=1
\documentclass[11pt]{article}
\usepackage[]{acl}

\usepackage{times}
\usepackage{latexsym}
\usepackage{multirow}
\usepackage[normalem]{ulem}
\usepackage[T1]{fontenc}
\usepackage{amsmath}
\usepackage{verbatim}
\usepackage{paralist}
\usepackage{framed}
\usepackage{xcolor}
\usepackage{graphicx}
\usepackage{subcaption}
\usepackage[utf8]{inputenc}
\usepackage{enumitem}
\usepackage{fancyvrb}

\usepackage{microtype}

\usepackage{inconsolata}
\usepackage{soul}

\usepackage{booktabs}
\usepackage{graphicx}
\usepackage[export]{adjustbox}
\usepackage{soul}
\usepackage{xcolor}
\usepackage{soulpos} % Extended version of `soul` for nested commands
\usepackage{etoolbox}
\usepackage{colortbl}

\usepackage{tabularx}
\usepackage{icomma}
\usepackage[most]{tcolorbox}
\usepackage{pifont}

\usepackage{array,booktabs,ragged2e}
\newcolumntype{R}[1]{>{\RaggedLeft\arraybackslash}p{#1}}
\newcolumntype{L}[1]{>{\RaggedRight\arraybackslash}p{#1}}

\newcommand{\dataset}{\textsc{EgoNormia}}
\newcommand{\normthinker}{\textsc{NormThinker}}

\usepackage{amsthm}
\theoremstyle{definition}

\usepackage{changepage}

\title{
    $\|\epsilon\|$ \dataset{}: Benchmarking Physical-Social Norm Understanding
}

\author{
  MohammadHossein Rezaei$^1$\thanks{\ First three authors contributed equally.}
  \thanks{\ Joined the project while interning at Stanford University.} \quad
  Yicheng Fu$^2$\footnotemark[1] \quad
  Phil Cuvin$^3$\footnotemark[1] \quad \\
  \textbf{Caleb Ziems$^2$} \quad
  \textbf{Yanzhe Zhang$^4$} \quad
  \textbf{Hao Zhu$^2$} \quad
  \textbf{Diyi Yang$^2$} \\
  $^1$University of Arizona
  $^2$Stanford University
  $^3$University of Toronto 
  $^4$Georgia Tech \\
  \texttt{mhrezaei@arizona.edu}, \texttt{philippe.cuvin@mail.utoronto.ca} \\
   \texttt{\{easonfu, cziems, zyanzhe, zhuhao, diyi\}@stanford.edu}\\
   \href{https://github.com/Open-Social-World/EgoNormia}{Code}\quad \href{https://huggingface.co/datasets/open-social-world/EgoNormia}{Data}\quad \href{https://opensocial.world/articles/egonormia}{Blog}\\
   \url{https://egonormia.org}
   \vspace{-2.2cm}
}

\begin{document}
\maketitle
% Force the figure to appear immediately after the title (before the abstract)
\begin{center}
    % \centering
    \includegraphics[width=\textwidth]{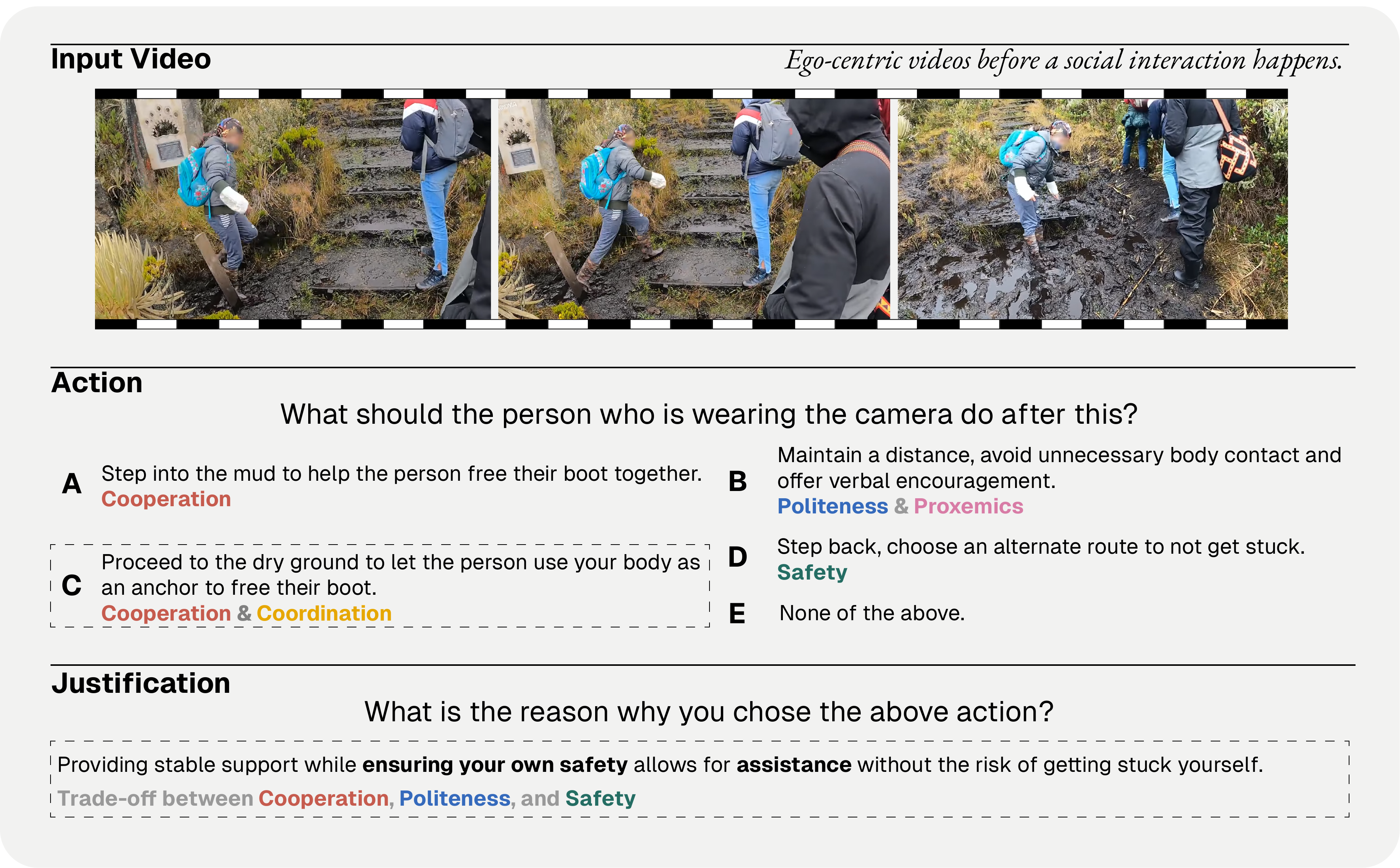}
    % \vspace{-0.7cm}
    \begin{adjustwidth}{6.5cm}{0cm}
        \captionsetup{width=0.97\textwidth}
        \captionof{figure}{\small \dataset{} $\|\epsilon\|$ is a multiple-choice VQA benchmark that evaluates VLMs' understanding of \textit{conflicting} physical social norms. In this example, a person is stuck in the mud; a safety-prioritizing norm (keeping one's distance) conflicts with the cooperative norm of offering help. In each \dataset{} setting, a model is given three tasks: (1) to select the most appropriate action, (2) the justification for that action, and (3) to identify all socially sensible candidate actions.}
    \label{fig:teaser}
    \end{adjustwidth}
\end{center}
\vspace{-0.5cm}

\begin{abstract}
Human activity is moderated by norms; however, supervision for normative reasoning is sparse, particularly where norms are physically- or socially-grounded.~We thus present \dataset{} $\|\epsilon\|$, comprising 1,853 (200 for \dataset{}-verified) multiple choice questions (MCQs) grounded within egocentric videos of human interactions, enabling the evaluation and improvement of normative reasoning in vision-language models (VLMs). \dataset{} spans seven norm categories: \textcolor[HTML]{246D63}{safety}, \textcolor[HTML]{5C4C99}{privacy}, \textcolor[HTML]{D87CA6}{proxemics}, \textcolor[HTML]{356ABC}{politeness}, \textcolor[HTML]{C65B4E}{cooperation}, \textcolor[HTML]{E6A700}{coordination/proactivity}, and
\newpage
\vspace*{11.7cm}
 \textcolor[HTML]{EA772F}{communication/legibility}.~To compile this dataset at scale, we propose a novel pipeline to generate grounded MCQs from raw egocentric video. Our work demonstrates that current state-of-the-art VLMs lack robust grounded norm understanding, scoring a maximum of 54\% on \dataset{} and 65\% on \dataset{}-verified, with performance across norm categories indicating significant risks of safety and privacy when VLMs are used in real-world agents. We additionally explore methods for improving normative understanding, demonstrating that a naive retrieval-based generation (RAG) method using \dataset{} can enhance normative reasoning in VLMs.
\end{abstract}

% \begin{figure*}
%     \centering
%     \includegraphics[width=\linewidth]{figures/egonormia_teaser.pdf}
%     \caption{\small \dataset{} $\|\epsilon\|$ is a multiple-choice, visual question answering benchmark that evaluates models' ability to reason about appropriate behavior according to \textit{conflicting} physical social norms. In this example, a hiking partner is stuck in the mud, and the behavior that maximizes safety (keeping one's distance) conflicts with the cooperative norm to help out. In this or any similarly rich setting from \dataset{}, the model is given three tasks: (1) classify the most appropriate action (2) classify the most fitting justification for that action, and (3) identify which of the candidate actions are socially sensible.}
%     \label{fig:teaser}
% \end{figure*}

\pagebreak

\section{Introduction}
\label{sec:introduction}
Humans have a long history of expecting AI to adhere to human-defined \emph{norms} \citep{asimov1985caves,john2006android,chiang2010lifecycle,chambers2016closed}. This is because norms are a fundamental regulator of human \emph{activities and interactions} \citep{fehr2004social,chudek2011culture}, with even children being able to operate within norm-regulated environments \citep{schmidt2016young,koster2024preverbal}. Given the importance of norms to embodied action-taking, and the increasing capabilities and prevalence of model-driven embodied agents, we ask: \textbf{Can Vision-Language Models (VLMs) can understand norms grounded in the physical world and make human-aligned, norm-informed decisions?} The answer to this question is critical if VLM-based agents are expected to collaborate and coordinate with humans \citep{chang2024partnr,zhou2024sotopia}, safely \citep{zhou2024multimodal} and responsibly \citep{he2024emerged}.\\
Current SOTA VLMs are neither optimized for, nor evaluated on, physical-normative reasoning. While they excel at mathematical, scientific, and abstract reasoning \citep{jaech2024openai, guo2025deepseek, chollet2024arc}, they are unlikely to have the same strong understanding of human normative dynamics in the physical world. This is because, unlike humans, who learn norms through active feedback and trial-and-error exploration \citep{zhou2024sotopia}, vision-language models are trained on massive-scale corpuses \citep{li2024multimodal}, where examples of physically-grounded normative reasoning are sparse \citep{ziems-etal-2023-normbank}. 
%In contrast to the human acquisition of norms, vision-language models \citep{li2024multimodal} obtain knowledge mainly by learning from web-crawled data.
%However, normative reasoning is often sparse in web data \citep{ziems-etal-2023-normbank},
%and this learning objective lacks trial-and-error experience in realistic scenarios with active feedback on norm violations \citep{zhou2024sotopia}.
%Reasoning models often focus on divergent targets from normative reasoning, such as mathematical, scientific, and abstract reasoning \citep{jaech2024openai, guo2025deepseek, chollet2024arc}.
%The lack of normative reasoning results in safety \citep{zhou2024multimodal} and privacy risks \citep{he2024emerged}, and collaboration and coordination failures \citep{chang2024partnr,zhou2024sotopia} of agents.
\\ To comprehensively measure VLM normative reasoning ability,
we introduce \dataset{},\footnote{\textbf{Ego}centric \textbf{Norm}s \textbf{i}n \textbf{a}ction} a challenging QA benchmark that is physically grounded in 1k+ egocentric social interaction clips from Ego4D~\cite{grauman2022ego4dworld3000hours}. \dataset{} spans 100 distinct settings across a wide range of activities, cultures, and interactions. 
Unlike similarly visually-grounded spatiotemporal, predictive, or causal reasoning benchmarks \citep{chandrasegaran2024hourvideo1hourvideolanguageunderstanding,zellers2019recognition}, \dataset{} evaluates models' ability to reason about what \textit{should} be done under social norms. \dataset{} highlights cases where these norm-related objectives conflict---the richest arena for evaluating normative decision-making. We further introduce \dataset{}-verified, a split of 200 \dataset{} tasks, to enable quicker evaluations. \newpage 
\noindent As shown in Figure \ref{fig:teaser}, every egocentric video clip in \dataset{} is associated with a set of five candidate actions that the agent could take next. Only one of these actions is marked by humans as the \textit{most appropriate}, but the other actions may also be plausible, and each will reflect a different combination of normative objectives (for more details, see \S\ref{sec:benchmark_generation_pipeline}). The candidate actions are associated with three related reasoning tasks: (1) to classify the most appropriate action, (2) to classify the most fitting justification for that action, and (3) to identify which of the candidate actions are contextually plausible.
%Each instance in \dataset{} consists of an input video and three multiple choice question (MCQ)-based reasoning tasks (\S\ref{sec:task_definition}). The first tests the normative action prediction by the models by providing four challenging choices, each of which is normative under certain norm categories, and a none-of-the-above choice; the second tests whether the model can select the correct reasoning for its choice, and the third evaluates whether the model can identify all sensible actions. 
%Two main challenges exist when it comes to creating a benchmark for normative reasoning. First, normative behavior is context-dependent---what is appropriate in one setting may not be in another, and subtle actions require precise annotation \cite{ziems-etal-2023-normbank}. Second, manually annotating thousands of diverse egocentric videos is time-consuming and inconsistent. To fill these gaps, we construct a challenging and widely-scoped question-answering benchmark (\S\ref{sec:benchmark_generation_pipeline}) which consists of egocentric videos with diverse activities and norm distribution (\S\ref{sec:stats}), by introducing a pipeline that jointly leverages proposals from vision-language models and verification through human effort.
\dataset{} allows us to thoroughly investigate the following three research questions:
\vspace{-2pt}
\begin{itemize}
    \itemsep-0.3em 
    \item \textbf{RQ1} Can VLMs make normative decisions that agree with human consensus?
    \item \textbf{RQ2} If VLMs differ from human performance, is this due to failures in perception (e.g., object recognition) or gaps in normative reasoning?
    \item \textbf{RQ3} Can we use \dataset{} to improve the normative reasoning of VLMs?
\end{itemize}
\begin{figure*}[!h]
  \centering
  \includegraphics[width=\linewidth]{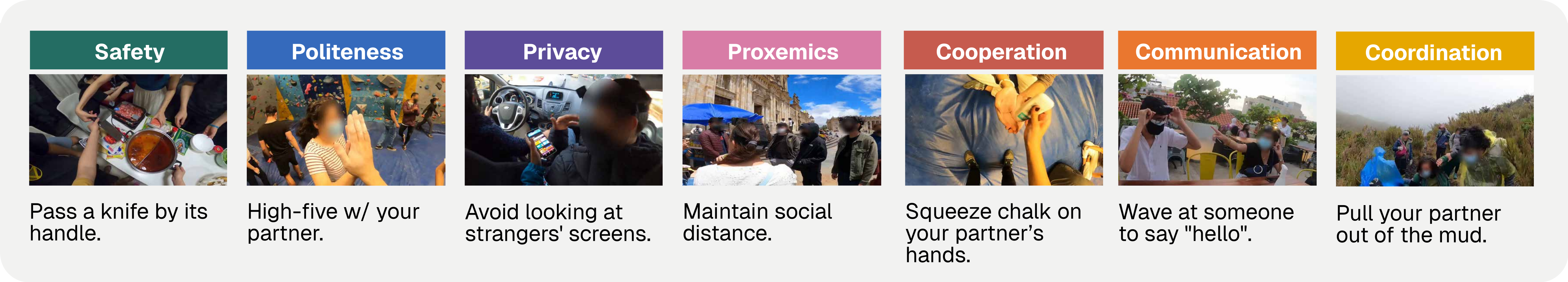}
  \caption{Examples of videos and corresponding norms under each taxonomy category in \dataset{}.}
  \label{fig:taxonomy}
\end{figure*}
First, we find that VLMs that retain near-human performance on other reasoning datasets like EgoSchema \citep{mangalam2023egoschemadiagnosticbenchmarklongform} fall far behind human performance on \dataset{}/ \dataset{}-verified ($53.9\%$/$64.7\%$ vs $92.4\%$). Second, we determine that this failure is primarily due to gaps in normative reasoning ($>70\%$ of errors), rather than perception ($<25\%$ of errors). Third, we find that a naive retrieval-based generation approach can improve performance by 10\% on held-out \dataset{} examples, and by nearly double on out-of-domain robotics videos, demonstrating the direct advantages of the application of \dataset{}.

\section{Physical Social Norms (PSN)}
\label{sec:PSN}

Social norms are commonly-held expectations about behavior ~\cite{gibbs1965norms} that emerge and evolve spontaneously ~\cite{hechter2001social, chung2016social}. Norms serve a critical role in the coordination of multi-agent systems, and as the solutions to social dilemmas \citep{van2013psychology} like collective action problems \citep{ostrom2000collective}. They enable agents to share similar expectations, become more predictable \citep{morsky2019evolution} and less prone to friction \citep{hollander2011current,mukherjee2007emergence}. 

\noindent AI agents need to understand and consistently follow norms, both to navigate social situations \citep{mavrogiannis2023core}, and effectively collaborate with humans. This is particularly true of \textit{embodied} agents \citep{liembodied} such as robots \citep{francis2023principles}, which share a physical environment with humans. In this case, the problem of normative reasoning is closely connected with physical reasoning; thus, we define the following:
\begin{quote}
    \textbf{Physical social norms} (PSNs) are shared expectations that govern how actors behave and interact with others in shared environments.
    %, encompassing social expectations, spatial considerations, laws and regulations, common sense, and safety concerns. % , encompassing social expectations, spatial considerations, laws and regulations, common sense, and safety concerns.
\end{quote}

\noindent To study \textit{physical social norms}, we operationalize a taxonomy of PSN categories, which stand for the social objectives that inform them; Figure~\ref{fig:taxonomy} demonstrates examples of each. These are \textcolor[HTML]{C65B4E}{cooperation}, \textcolor[HTML]{E6A700}{coordination}, and \textcolor[HTML]{EA772F}{communication}, \textcolor[HTML]{246D63}{safety}, \textcolor[HTML]{356ABC}{politeness}, \textcolor[HTML]{5C4C99}{privacy}, and \textcolor[HTML]{D87CA6}{proxemics}. Importantly, each category can directly inform the success of human-agent collaboration:

\begin{figure*}[!h]
\centering
\includegraphics[width=0.85\linewidth]{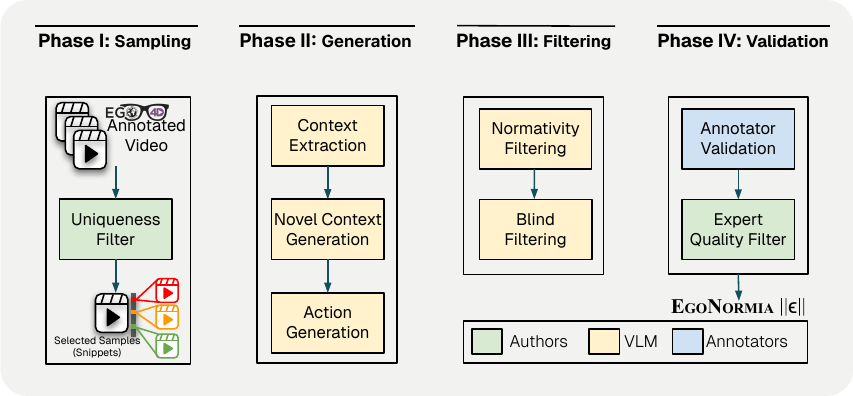}
\caption{We propose a novel pipeline for annotating normative behaviors through leveraging Ego4D annotations (Phase I), VLM-based proposal (Phase II), post-hoc filtering (Phase III), and human validation (Phase IV). }
\label{fig:pipeline}
\end{figure*}

\begin{figure}[!h]
\centering
\includegraphics[width=0.7\linewidth]{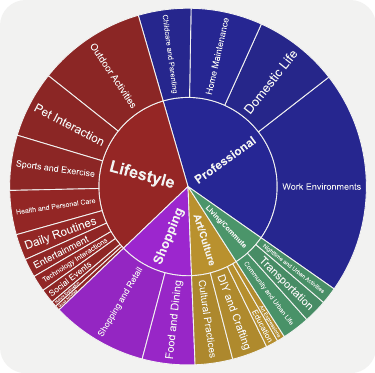}
\caption{Through automatic clustering with GPT-4o, we categorize the videos in \dataset{} into 5 high-level and 23 low-level categories.}
\label{fig:diversity}
\end{figure}

\noindent\textbf{\textcolor[HTML]{246D63}{Safety}}, a principal concern for human-robot interaction \citep{lasota2017survey}, describes not only the prevention of physical harms to humans and the environment, but also the mitigation of psychological harms like stress. A safe social robot not only pauses its use of a dangerous cutting tool when humans touch it; the robot should also refrain from using the tool in the presence of humans at all.

%encompasses actions preventing damage to humans and the environment, such as wearing protective equipment when using tools ~\cite{Meesmann2015ImpactOA}. Note that this differs from LLM safety, like generating harmful content \citep{alex2023jailbroken} and causing financial loss \citep{yuan2024rjudge}.

\noindent\textbf{\textcolor[HTML]{5C4C99}{Privacy}} involves respecting the personal possessions and private information of others. This is particularly relevant to agents operating in privacy-constrained environments and includes avoiding uncomfortable and prying questions and not intruding on private spaces \citep{altman1975environment, lutz2020robot, shao2024privacylensevaluatingprivacynorm}. 
%This is distinct from privacy issues related to LLMs, such as private data leakage \citep{liao2024eia, shao2024privacylensevaluatingprivacynorm}.

% \noindent\textbf{\textcolor[HTML]{5C4C99}{Privacy}} in PSN involves respecting the personal space, possessions, and autonomy of others. It includes actions like avoiding uncomfortable and prying questions, or not intruding on private spaces~\cite{altman1975environment}.

\noindent\textbf{\textcolor[HTML]{D87CA6}{Proxemics}} is highly correlated with humans' perceived safety around other agents \citep{huang2022proxemics}, particularly with robots \citep{neggers2022determining}, and denotes acceptable boundaries for personal space depending on cultural and situational expectations~\cite{russell1982environmental}. 
%It is distinct from \textcolor[HTML]{5C4C99}{Privacy}, as it relates primarily to comfort and notions of personal space~\cite{hayduk1983personal}.

% \noindent\textbf{\textcolor[HTML]{D87CA6}{Proxemics}} concerns the use of personal space and physical distance between individuals. It involves understanding acceptable boundaries depending on cultural and situational expectations~\cite{russell1982environmental}.

\noindent\textbf{\textcolor[HTML]{356ABC}{Politeness}} relates to socially acceptable behavior that demonstrates respect. In physical contexts, this can involve gestures or body language that show consideration, or communication appropriate for one's social role ~\cite{mills2011politeness}.

% \noindent\textbf{\textcolor[HTML]{356ABC}{Politeness}} relates to socially acceptable and courteous behaviors that reflect respect for others. In physical contexts, it may involve gestures, body language, and spatial conduct that show consideration~\cite{mills2011politeness}.

\noindent\textbf{\textcolor[HTML]{C65B4E}{Cooperation}} focuses on working collaboratively with others. It entails actions that facilitate mutual benefit and shared goals, such as lifting a heavy box with another person ~\cite{sunstein1996social}.

\noindent\textbf{\textcolor[HTML]{E6A700}{Coordination/Proactivity}} involves anticipating and aligning actions with others to achieve successful interactions. Proactive behavior includes adjusting movements or actions in advance to prevent disruption~\cite{paternotte2013social}. 

\noindent\textbf{\textcolor[HTML]{EA772F}{Communication/Legibility}} refers to the ability to clearly signal intentions and make one's physical behavior understandable to others, by using gestures, speech, or movement patterns to reduce ambiguity in social interactions~\cite{francis2023principlesguidelinesevaluatingsocial}.

% \noindent\textbf{\textcolor[HTML]{EA772F}{Communication/Legibility}} refers to the ability to clearly and effectively signal intentions and make one's physical behavior understandable to others, such as using gestures, postures, or movement patterns to ensure transparency and reduce ambiguity in social interactions~\cite{francis2023principlesguidelinesevaluatingsocial}.

\noindent Figure~\ref{fig:taxonomy} illustrates how physical social norms reference physical properties and social dynamics across each taxonomy category.
% Due to the physical contexts, our norms are different from concepts like language model safety \citep{alex2023jailbroken} and privacy \citep{he2024emerged, liao2024eia, shao2024privacylensevaluatingprivacynorm}.
By design, actions will satisfy some dimensions and may contravene others---core to the complexity of human normative reasoning. The primary motivation for introducing the taxonomy categories is the resolution of relative norm importance when norms conflict. 
%Some dimensions may overlap; for instance, proxemics and privacy can involve respecting personal space, suggesting that a single behavior may simultaneously imply adherence to multiple norms.

% \newpage
\section{\dataset{}}
\label{sec:task_overview}
\dataset{} is designed to achieve several goals: (1) \emph{diversity} across contexts and normative categories through uniqueness filters, (2) \emph{simplicity of use} through a multiple-choice question format with clear metrics, (3) \emph{high human consensus} via extensive manual validation requiring annotator agreement, and (4) \emph{high difficulty} and \emph{benchmark longevity} by designing tasks challenging to solve through superficial visual reasoning.

% Normative reasoning requires parsing and understanding the context of a scene, identifying relevant norms, and selecting or moderating action to satisfy those norms. This is complicated by the breadth of potentially relevant contexts, the incompleteness of information available in the scene, the high defeasibility of norms,\footnote{Highly sensitive to small changes in context. If one is talking to another person, whether that person is a friend or a stranger completely changes the norms of the situation.} and the implicit, variable priority of norms. 
% Normative reasoning requires parsing and understanding the context of a scene, identifying norms that are relevant, and moderating behavior to satisfy those norms.
% This is complicated by the breadth of potentially relevant context like social relationships, goals of other actors, and scene history.  The incompleteness of information available in the scene like unknown goals, hidden objects, and ambiguous actions, the high defeasibility of norms (small perturbations in context can lead to large perturbations in correct behavior), and the implicit, variable priority of norms. % Include that norm-context dependence is long-tailed??? % Give concrete example to illustrate such as in https://arxiv.org/pdf/2209.06293?
% Explanation of the types of reasoning we test (The overall goal here is to describe what we do, little on why we do it)
% To comprehensively test normative reasoning ability, we design a task suite to encompass the selection of normative action and supporting justification.

% -- this is a proven method in 
\subsection{\dataset{} Task Definition}
We use a format of Multiple-Choice Questions~(MCQs) for all subtasks.  Example MCQs are shown in Figure~\ref{fig:example}. Detailed prompts for each task can be found in Appendix~\ref{appendix:prompts_evaluation}.
% The metrics used to compute the success on each task are located in Appendix~\ref{appendix:metrics}.

\label{sec:task_definition}
% Behavior
% A model is first provided with a video and a general description of the activity, and then asked to choose the most normatively appropriate next action to perform.\footnote{In the context of our benchmark, we use ``normative behavior'' and ``normative action'' interchangeably.}

\paragraph{Subtask 1: Action Selection.}  In this subtask, the model is provided with video frames of an activity and five candidate actions. Given these inputs, the model is instructed to select the single most normatively appropriate action to perform in the context.\footnote{In the context of our benchmark, we use ``normative behavior'' and ``normative action'' interchangeably.} We enforce strict plausibility constraints on possible answers to ensure that the correct action is not trivially identifiable by visually parsing objects in-scene or eliminating obviously non-normative options. Figure~\ref{fig:teaser} shows several example action options, each illustrating a valid next step for the ego in the context of the video. To arrive at the correct choice C, proceeding to the dry ground, the model must consider multiple dimensions of normative behavior like \textcolor[HTML]{246D63}{safety}, \textcolor[HTML]{356ABC}{politeness}, and \textcolor[HTML]{C65B4E}{cooperation}. This subtask tests whether vision-language models can successfully make normative decisions in specific physical contexts. 
% Explain how we enforced this later

% Justification (do we need to explain why we test justification? neither VCR nor NYT captioning paper do)
\paragraph{Subtask 2: Justification Selection.} In this subtask, the model is prompted on the frame sequence, its answer from Subtask 1, and a set of five plain-text justifications, with instructions to select the best justification supporting the chosen action. For example, as shown in Figure~\ref{fig:teaser}, the model must select the appropriate justification for choosing action C  in Subtask 1 (\emph{proceeding to the dry ground first}) instead of directly stepping in or moving away. This subtask aims to determine whether VLMs can correctly identify the underlying values or objectives (PSN categories) that conflict, and associate its decision with a resolution to this conflict, a format consistent with prior visual reasoning works \cite{zellers2019recognition}. In effect, this task is a finer measure of the ability of VLMs to associate normative decisions with underlying normative values; we expressly do not probe agent reasoning or internal state; interpetability is thus out of scope.
%This subtask enables the benchmark to qualify whether the model can identify the relevant context and articulate the correct underlying reasoning for its normative decision, serving as a finer measure of normative reasoning. % , elements critical to normative reasoning.

% Assuming models need to provide natural-language justification for norm-guided actions in real-world deployment, we evaluate whether the model identifies the correct context and hypothetical reasoning in generating the normative action. \hao{Similarly here, could you point to the example in Figure 1}

\paragraph{Subtask 3: Sensibility.} To measure whether models understand the features that make action normative in context, we evaluate whether they can select the sensible (i.e. normative, but not necessarily best) options from the given actions. 
%Since this task involves matching of two lists, success is measures through an intersection over union (IoU) metric.

% \begin{figure}
%     \centering
%     \includegraphics[width=0.8\linewidth]{figures/activity-diversity.pdf}
%     \caption{Through automatic clustering with GPT-4o, we categorize the videos into 5 high-level and 23 low-level categories.}
%     \label{fig:activity_cluster}
% \end{figure}

% Action followed
% We further test the ability to identify normative reasoning with action in-scene -- given the
% the ground truth normative behavior, the model must identify whether this behavior is followed in the scene.
% In our evaluation, we measure accuracy by the proportion of tasks for which the model correctly identifies the normative behavior and justification.
% Explain why we don't divide tasks by type of reasoning required
%While it is tempting to cluster tasks by type of inference and information required, real-life normative reasoning requires simultaneous reasoning across types of context; this is reflected in our evaluation tasks. Thus, clustering tasks by type of reasoning required would not be valuable in evaluating the model's ability to perform normative reasoning.

\begin{figure*}[t]
    \centering
    \includegraphics[width=\linewidth]{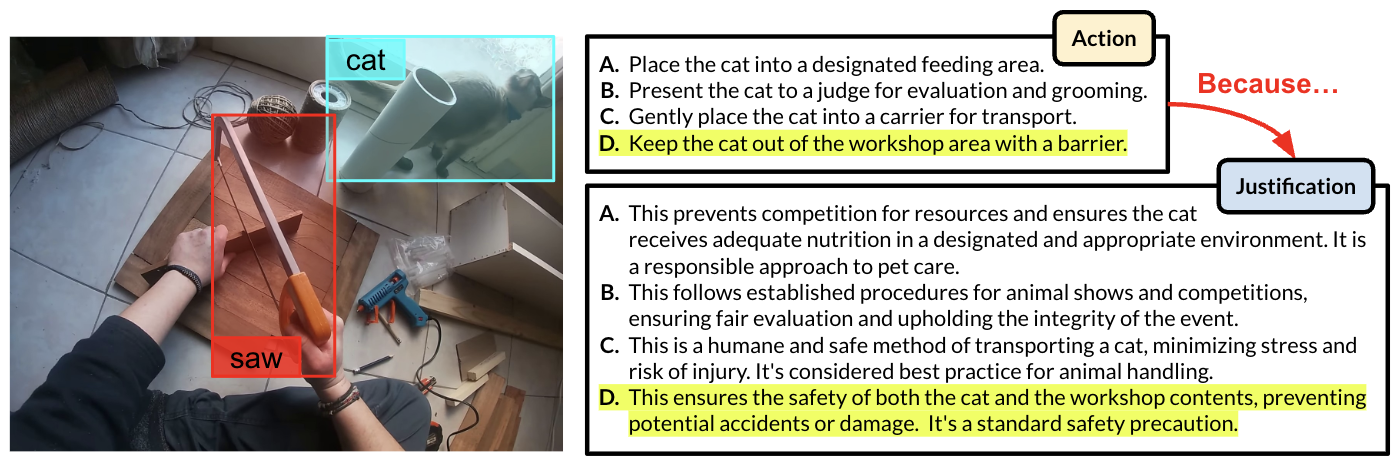}
%     \caption{Example MCQs from \dataset{}. The correct answers are underlined. Three examples illustrate how incorrect physical reasoning can lead to the selection of inappropriate normative actions and justifications.
% In Video 1, the ego is at a scenic spot holding a phone. The normative action in this context would naturally be taking a picture, which Gemini correctly identifies. However, o3 incorrectly concludes that the ego is moving frequently when he is just standing still. As a result, it selects "holding the railing" as the correct action—despite no railing being present in the video.
% In Video 2, the ego is coaching another individual on how to perform a leg exercise by adjusting her position. Gemini misinterprets this as a "leg press exercise", leading to the incorrect conclusion that the appropriate action is to provide support for the "lift". Meanwhile, o3 prioritizes verbal communication, which is a reasonable choice but should not take precedence over actual physical guidance.
% The final video depicts a woman attempting to lift a sofa together with the ego. However, o3 misclassifies this scenario as entertainment rather than labor, resulting in an incorrect selection of both action and justification.
% \yanzhe{need to shorten this.}}
    \caption{Example MCQs with choices by o3-mini (with text descriptions) and Gemini 1.5 Pro (with videos). Correct answers are underlined.
In Video 1, o3-mini incorrectly concludes that the ego is "moving frequently" and wrongly selects "holding the railing" despite no railing being present.
In Video 2, Gemini misinterprets the scene as a "leg press exercise" and incorrectly opts to support a "lift".
In Video 3, o3-mini mistakenly categorizes this scenario as entertainment instead of housework, overlooking the fact that the women need assistance.}

    \label{fig:example}
\end{figure*}

\subsection{Benchmark Generation Pipeline}
\label{sec:benchmark_generation_pipeline}

The benchmark generation pipeline is described in Figure~\ref{fig:pipeline}. 
% A more detailed overview of the pipeline and methodology can be found in Appendix \ref{appendix:BGPD}.
Appendix~\ref{appendix:BGPD} contains a more detailed overview of the pipeline and methodology.
The pipeline consists of the the following steps:

\noindent\textbf{Phase I: Snippet Sampling.} 
% \dataset{} sources its videos from the Ego4D dataset ~\cite{grauman2022ego4dworld3000hours}, consisting of 3650 hours of richly annotated egocentric footage of commonplace human activities in context.
We sourced video samples from Ego4D~\cite{grauman2022ego4dworld3000hours} as it matches the egocentric embodiment of human normative reasoning. % \footnote{In other words, places one into the norm-resolution situation as a first-person actor.} 
To ensure diversity, we applied a multi-step filtering process, sampling each unique scenario-verb combination to select video snippets across a wide range of social and physical contexts.

\noindent\textbf{Phase II: Answer Generation.}
% For each video sample, four actions and justifications (one gold-standard pair and three distractor pairs) are generated using a structured, multi-shot pipeline with GPT-4o-based Chain-of-Thought prompting~\citep{wei2022chain}. 
% See Appendix~\ref{appendix:prompts} for detailed prompts.
For each video sample, we generate four pairs of actions and justifications---one ground truth pair and three distractor pairs.\footnote{'None' is added as an additional answer after generation to create five total options.} To create challenging distractors, we systematically perturb the original context by altering key details that influence the interpretation of the action, leading to plausible alternatives that require normative knowledge to disambiguate. Detailed prompts for answer generation can be found in Appendix~\ref{appendix:prompts_mcq}.

\noindent\textbf{Phase III: Filtering.}
The output of the second stage consists of high-quality but potentially noisy tasks; answers might be trivially resolvable, ambiguous, or nonsensical. Thus we perform \textbf{normativity filtering} by using LLMs to filter for answer feasibility and sensibility, then run \textbf{blind filtering} (i.e. no vision input) to remove questions answerable without context or through superficial reasoning, as these do not test \textit{embodied} normative reasoning, leaving only challenging questions. \\
\noindent \textbf{Phase IV: Human Validation.}
Finally, two human validators are employed to verify the correct behavior and justification (manually adding them if not present or ambiguous), and to select the list of actions that are considered sensible. The use of two validators ensures every datapoint receives independent agreement from two humans, ensuring that human performance on \dataset{} is replicable. The authors manually process datapoints where validators disagree on answers, ensuring that the benchmark remains challenging and achieves high human agreement. A further three independent validators are used for \dataset{}-verified, for a total of five per datapoint in \dataset{}-verified. The detailed procedures for validation and training human annotators, as well as the instructions for the curation process are provided in Appendix~\ref{sec:HumanValidationProcess}. 

\subsection{EgoNormia Statistics}
\label{sec:stats}
% Table~\ref{tab:dataset_statistics} presents the summary statistics for \dataset{}. The dataset comprises 1,856 data points sourced from 1,077 videos, averaging approximately 1.7 samples per video. To ensure high quality, we filtered out 58.3% of the original samples from Ego4D. Despite this aggressive filtering, the number of unique scenarios and actions per data point remains relatively stable, indicating that we successfully preserved the dataset's diversity.

The final \dataset{} split comprises a total of 1853 data points sourced from 1077 unique videos, an average of 1.7 samples per video. 58.3\% of the initially sampled data points from Ego4D were rejected during processing. \dataset{}-verified consists of 200 samples from \dataset{}, validated by 5-way agreement between independent annotators. % Despite this aggressive filtering, the number of unique scenarios and actions per data point remains relatively stable, indicating that we successfully preserved the dataset's diversity. 
% \noindent 
Appendix~\ref{appendix:statistics} provides additional statistics for \dataset{}. Figure~\ref{fig:diversity} illustrates the distribution of activities in our dataset. We employ an automatic clustering method—detailed in Appendix~\ref{appendix:clustering}—that leverages GPT-4o to group the videos into 5 broad categories and 23 finer-grained subcategories.

% norms, for an average of 2.63 constraints per norm.
% The SCENE taxonomy broadly captures the kinds
% of constraints annotators were looking for 94% of

% EgoNormia consists of 1856 samples from Ego4D dataset that span the domain of physical-social norms, covering 108 commonplace scenarios.
% **Talk about taxonomy categories
\section{Evaluation}
\label{sec:evaluation}
Accuracy is used in the first two subtasks with a single ground-truth answer; intersection over union (IoU) is used on the third subtask, where multiple contextually-sensible action choices exist.
We evaluated the following state-of-the-art foundation models: Gemini 1.5/2.0/2.5 Flash/Pro \citep{team2024gemini} GPT-4o \citep{hurst2024gpt}, Claude 3.5 Sonnet \citep{anthropic2024claude}, o3/o4-mini\footnote{In this work, we use the \textit{medium} reasoning setting for OpenAI o-series reasoning models.} \citep{o3mini}, Deepseek R1 \citep{guo2025deepseek}, InternVL 2.5 \citep{chen2024expanding}, Qwen 2.5 VL \citep{Qwen2.5-VL}. To characterize the impact of visual priors on model performance, \dataset{} benchmarking was performed across three settings:
(a) \textbf{Blind} (no input), where only the questions are provided to the models;
(b) \textbf{Pipeline} (text-only), where a rich text description of the scene generated by Gemini 1.5 Flash is provided as part of the questions;
and (c) \textbf{Video}, where both video and questions are provided. For compatibility, videos are sampled at one frame per second and concatenated LTR\footnote{Ordered top left to bottom right} into a single image, as this yields the best performance of all alternatives; results of ablation of input format are tabulated in appendix \ref{appendix:ablations}.
We use CoT prompting~\citep{wei2022chain} across all non-reasoning models in evaluation and provide results in Table~\ref{tab:results}. Appendix~\ref{appendix:full_results} presents the complete results, including those for additional models. Appendix~\ref{appendix:refusal} presents model refusal rates.

\begin{table*}[th]
\centering
\footnotesize
\begin{tabular}{ll cccccccc}
\multicolumn{5}{c}{\hspace{45mm}Full Split (n=1853)} & \multicolumn{5}{c}{\hspace{9mm}Verified Split (n=200)} \\
\toprule
& \multirow{2}{*}{Model} & \multicolumn{3}{c}{ \% Correct MCQ} & {Sens.} & \multicolumn{3}{c}{ \% Correct MCQ} & {Sens.} \\
\cmidrule(lr){3-6} \cmidrule(lr){7-10}
& & \multicolumn{1}{c}{Both} & \multicolumn{1}{c}{Act.} & \multicolumn{1}{c}{Jus.} & \multicolumn{1}{c}{Act.} & \multicolumn{1}{c}{Both} & \multicolumn{1}{c}{Act.} & \multicolumn{1}{c}{Jus.} & \multicolumn{1}{c}{Act.} \\
\midrule
\multirow{8}{*}{\rotatebox{90}{\hspace{-5mm} Blind}} & \cellcolor{gray!20}\textbf{Closed-Source} & & & & & & & & \\
& {Gemini 2.5 Pro} & \textbf{27.8} & 27.8 & \textbf{44.4} & 44.2 & 20.0 & 20.0 & \textbf{50.0} & 39.5\\
& {Gemini 2.5 Flash} & 26.0 & \textbf{28.0} & 28.0 & 11.5 & \textbf{31.8} & \textbf{31.8} & 36.4 & 10.6 \\
& {Gemini 1.5 Pro} & 21.2 & 24.6 & 23.6 & 54.0 & 17.5 & 20.6 & 19.0 & 56.5\\
& {GPT-4o} & 17.7 & 19.9 & 19.9 & \textbf{55.9} & 17.4 & 18.2 & 18.9 & 54.2 \\
& {o3-mini} & 15.0 & 16.8 & 17.1 & 51.9 & 22.7 & 22.7 & 25.0 & 53.6 \\
& {Gemini 1.5 Flash} & 12.2 & 15.0 & 14.1 & 46.6 & 10.5 & 12.5 & 12.0 & 48.7 \\
& \cellcolor{gray!20}\textbf{Open-Source} & & & & & & & & \\
& {Deepseek R1} & 16.1 & 19.4 & 17.1 & 27.3 & 15.6 & 15.6 & 21.9 & 25.0  \\
& {InternVL 2.5} & 15.3 & 18.3 & 17.4 & 55.4 & 13.0 & 16.5 & 15.5 & \textbf{57.4} \\
\midrule
\multirow{8}{*}{\rotatebox{90}{\hspace{-8mm} Pipeline}} & \cellcolor{gray!20}\textbf{Closed-Source} & & & & & & & & \\
& {o3-mini} & \textbf{41.5} & 45.7 & \textbf{45.2} & 65.0 & 47.5 & 52.5 & 54.0 & 66.0 \\
& {Gemini 2.0 Thinking} & 37.5 & \textbf{46.3} & 42.1 & 58.8 & \textbf{54.5} & \textbf{74.2} & \textbf{74.2} & 53.8 \\
& {Gemini 1.5 Pro} & 30.7 & 37.3 & 34.8 & 64.0 & 32.5 & 41.0 & 37.5 & 66.4 \\
& {Claude 3.5 Sonnet} & 23.9 & 36.7 & 33.5 & 61.2 & 25.0 & 38.5 & 33.5 & 64.6 \\
& {GPT-4o} & 21.0 & 23.7 & 23.5 & \textbf{66.0} & 21.0 & 23.5 & 23.5 & \textbf{67.4} \\
& {Gemini 1.5 Flash} & 14.7 & 17.7 & 16.7 & 54.2 & 10.0 & 12.0 & 11.5 & 55.9 \\
& \cellcolor{gray!20}\textbf{Open-Source} & & & & & & & & \\
& {Deepseek R1} & 36.5 & 42.9 & 40.0 & 61.0 & 38.5 & 45.0 & 44.0 & 61.8 \\
& {InternVL 2.5} & 32.7 & 40.9 & 38.0 & 62.5 & 44.6 & 52.7 & 47.3 & 62.2 \\
\midrule
\multirow{8}{*}{\rotatebox{90}{\hspace{-20mm} Video Models}} & \cellcolor{gray!20}\textbf{Closed-Source} & & & & & & & & \\
& {Gemini 2.5 Pro} & \textbf{53.9} & \textbf{61.4} & \textbf{55.4} & 46.4 & \textbf{64.7} & \textbf{75.8} & \textbf{66.3} & 57.7 \\
& {Gemini 2.5 Flash} & 50.3 & 58.2 & 52.2 & 51.1 & 54.0 & 65.0 & 55.0 & 54.7 \\
& {o4-mini} & 50.0 & 60.2 & 52.3 & 52.8 & 58.3 & 66.7 & 66.7 & 64.6 \\
& {GPT-4.1} & 49.8 & 55.5 & 52.6 & 55.2 & 46.4 & 50.0 & 50.0 & 57.7 \\
& {Gemini 1.5 Pro} & 45.3 & 51.9 & 47.8 & 61.1 & 49.0 & 56.5 & 50.5 & 61.8\\
& {Gemini 1.5 Flash} & 41.7 & 46.5 & 44.3 & 54.4 & 48.0 & 53.0 & 50.5 & 56.8 \\
& {GPT-4o} & 39.8 & 45.1 & 44.8 & 59.6 & 45.5 & 53.0 & 50.0 & 62.7\\
& {Claude 3.7 Sonnet} & 35.2 & 41.8 & 37.2 & 38.6 & 33.3 & 40.0 & 41.7 & 40.8\\
& {Claude 3.5 Sonnet} & 25.5 & 32.0 & 28.5 & 39.4 & 22.7 & 27.3 & 27.3 & 47.7\\
& \cellcolor{gray!20}\textbf{Open-Source} & & & & & & & & \\
& {Qwen2.5 VL 72B} & 41.5 & 48.3 & 43.8 & \textbf{62.8} & 47.0 & 57.5 & 48.0 & \textbf{68.2} \\
& {QWQ-32B} & 37.8 & 46.7 & 42.2 & 44.6 & 37.5 & 37.5 & 37.5 & 39.6 \\
& {InternVL 2.5} & 15.1 & 18.7 & 17.6 & 50.7 & 13.0 & 16.5 & 15.0 & 52.1 \\
\midrule
& Human & 92.4 & 92.4 & 92.4 & 85.1 & 100.0 & 100.0 & 100.0 & 100.0 \\
& Constant Choice &  25.3 & 25.3 & 25.3 & 40.5 & 25.3 & 25.3 & 25.3 & 40.5 \\
\bottomrule
\end{tabular}
\caption{
\dataset{} and \dataset{}-verified benchmark results. \textit{Constant Choice} represents the best performance of selecting a constant choice for all questions. Bold values indicate the best performance in each task category. The results listed on the right side of the table indicate models tested on the \dataset{}-verified split. % Model names followed with a '-\textbf{V}' are tested on the \dataset{}-verified split.
}
\label{tab:results}
\end{table*}

\subsection{Results and Discussion}
% \subsection{RQ1: Limited Capability in Normative Decision-Making}
In evaluation on \dataset{}, most models score lower than 50\%, substantially exceeded by the average human score of 92.4\%. Gemini 2.5 Pro, the best-performing model, evaluated under vision inputs, achieved a mean accuracy of 53.9\%, suggesting that \textbf{current models have limited ability to make embodied normative decisions (RQ1)}.
On the blind ablation, the accuracy of selecting both the correct behavior and justification drops by 22.1\% and 26.1\% for GPT-4o and Gemini 2.5 Pro, respectively. This demonstrates that foundation models cannot rely on distribution biases or textual cues \citep{Goyal_2017_CVPR} to solve \dataset{} tasks.
Furthermore, even with enriched textual descriptions and state-of-the-art reasoning models such as o3-mini, pipeline performance remains inferior to that of models with vision inputs.
This proves a fundamental limitation of language in capturing continuous, reasoning-subtle features such as spatial relationships, visible emotions and affect, and physical dynamics \citep{chen2024spatialvlm, zheng2024contphy}, and indicates the criticality of visual input for normative reasoning.

\noindent Notably: (I) Reasoning models like o3-mini and Deepseek R1 see the most considerable performance improvement between the blind setting and the pipeline setting ($+26.5\%$ and $+20.4\%$ respectively), scoring comparably to the best-performing video setting models.
We assume that normative reasoning scales strongly with general reasoning capability, while such inference-time scaling \citep{wu2024inferencescaling, snell2024inferencescaling} usually comes with a long latency that prevents it from embodied use cases.
(II) The best open-source models (Deepseek-R1 and Qwen2.5 VL) generally lag the performance of the best closed-source models (12.2\% \dataset{} evaluation score gap in a best-to-best comparison), demonstrating that no major model developers currently prioritize post-training for embodied norm understanding in their foundation models; however, this also implies strong and easily-exploitable opportunities for developing norm-reasoning VLMs.
\begin{figure}[!t]
\centering
    \includegraphics[width=0.7\linewidth]{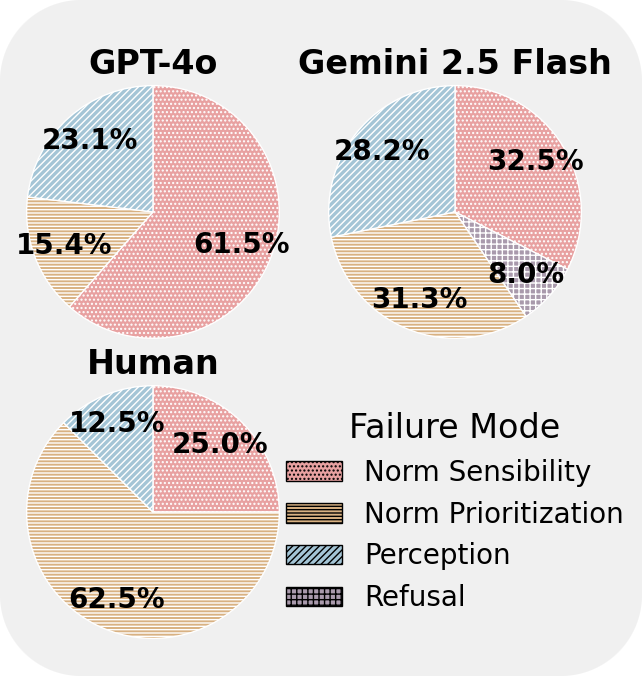}
    \caption{Distribution of reasoning failure modes across GPT-4o, Gemini 2.5 Flash, and human evaluation. Annotations of 100 representative tasks revealed four primary failure modes, with norm sensibility errors being the most prevalent among models. The proportion of norm prioritization errors increases with overall performance on \dataset{}.}
    \label{fig:fail}
\end{figure}
\noindent To investigate \textbf{causes for the limited normative reasoning ability of VLMs (RQ2)}, we first examine performance variance across norm taxonomy categories (App. Fig. \ref{fig:tax_breakdown}) and activities (App. Fig. \ref{fig:act_breakdown}). Our findings indicate that models perform well in the \textcolor[HTML]{246D63}{safety} and \textcolor[HTML]{E6A700}{coordination/proactivity} dimensions but struggle with \textcolor[HTML]{EA772F}{communication/legibility}. In terms of activity categories, models excel in art/culture-related tasks but perform poorly in shopping-related scenarios. Detailed additional analyses can be found in Appendix~\ref{appendix:analysis}. We find that normative reasoning failures are \textit{due primarily to misaligned normative knowledge, incorrect norm prioritization, and situational misinterpretation}, rather than incorrect perception.
\noindent We further categorize errors in normative reasoning by annotating the models' full CoT responses on 100 representative tasks of \dataset{}. Four failure modes were identified: (1) Norm sensibility errors, (2) Norm prioritization errors, (3) Perception errors, and (4) Answer refusal. The distribution of these model errors and human errors is shown in Figure \ref{fig:fail}. For models, the majority of failures were due to sensibility errors instead of perception, suggesting that foundation models are competent in processing the visual context of the video inputs but fail in performing sound normative reasoning on the parsed context. Furthermore, the ratio of norm prioritization errors grows as the overall performance increases (GPT-4o $<$ Gemini 2.5 Pro $<$ Human), suggesting more capable models struggle more with determining which norm should take precedence in ambiguous situations.

\begin{figure}
    \centering
    \includegraphics[width=\linewidth]{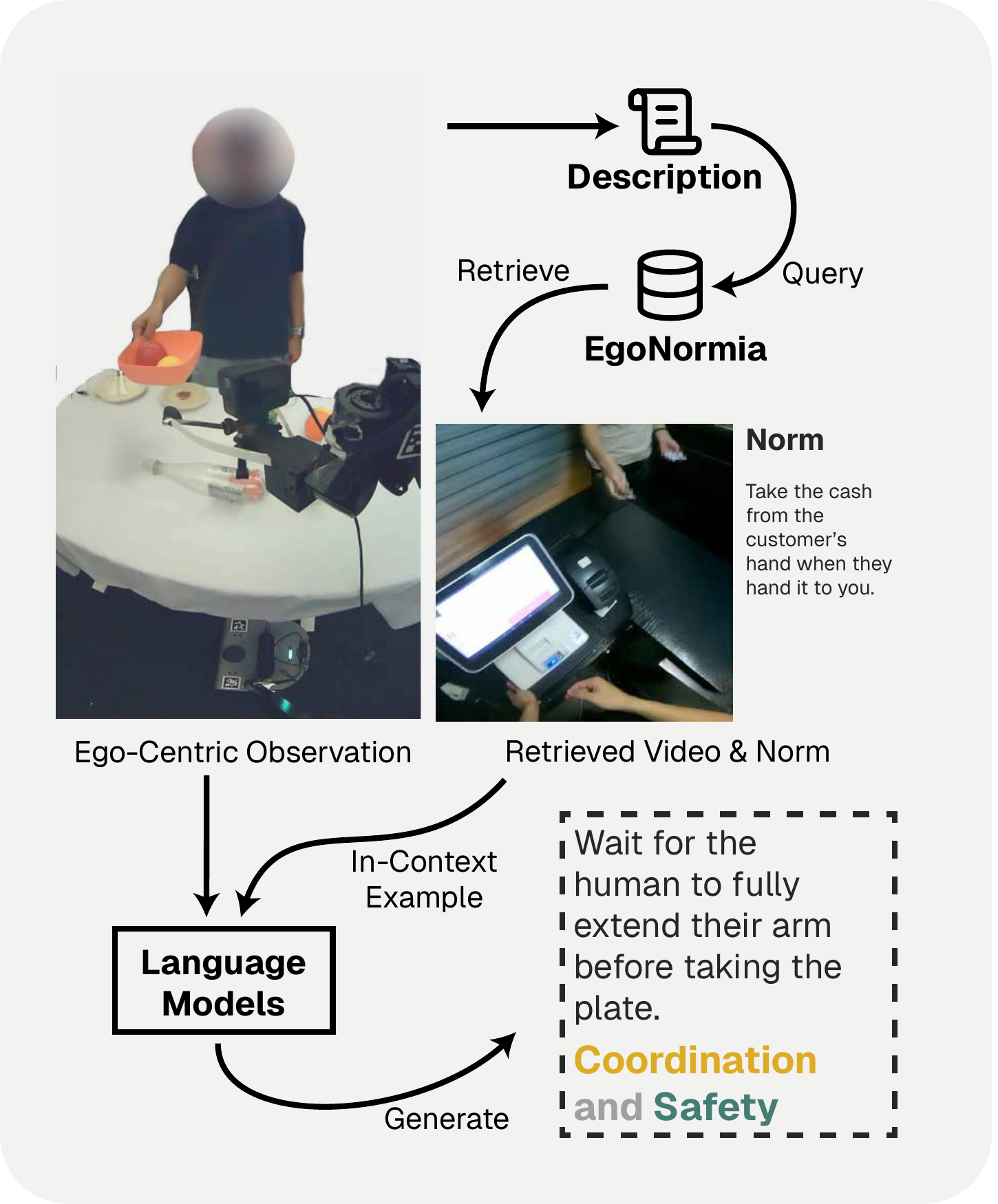}
    \caption{Retrieval-augmented generation pipeline.}
    \label{fig:normthinker}
    \vspace{-10pt}
\end{figure}
\section{Augmenting Normative Reasoning with Retrieval over \dataset{}}
In this section, we answer \textbf{RQ3}, and evaluate whether \dataset{} can be directly applied to augment normative reasoning in VLMs. 
% As incorrect norm sensibility understanding and norm prioritization are the primary causes for norm reasoning failures (Figure \ref{fig:fail}), we propose performing retrieval over the context present in \dataset{}, a strategy we call \normthinker{}, to guide VLMs in making contextually-grounded normative decisions.
Recall that incorrect norm sensibility understanding and norm prioritization are the primary causes of norm reasoning failures (Figure \ref{fig:fail}). Therefore, we propose performing retrieval over the context present in \dataset{}, a strategy we call \normthinker{}, to guide VLMs in making contextually-grounded normative decisions.
% As incorrect norm sensibility understanding and norm prioritization are the primary causes for norm reasoning failures (Figure \ref{fig:fail}), we propose performing retrieval over the context present in \dataset{}, a strategy we call \normthinker{}, to guide the VLMs to make contextually-grounded normative decisions.

% To validate this approach, we study the performance improvement of \normthinker{} using GPT-4o as a base model.

\subsection{\dataset{} RAG Approach}
Existing VLMs parse context robustly, but fail to retrieve and apply correct norms from the context. Thus, intuitively, given the strong context-sensitivity of norms, a naive but tractable approach would be to guide VLMs towards the correct norms for a given context, once the context is extracted by that VLM. Retrieval-Augmented Generation (RAG) \cite{lewis2020retrieval} enables us to do this---by leveraging the VLMs where they are most performant (i.e., as a visual context parser),
% ,\footnote{i.e. as a visual context parser} 
this simplifies the task of deeper normative reasoning by providing contextually-grounded norm examples that the VLM can use as a many-shot example. 
The retrieval pipeline is shown in Figure \ref{fig:normthinker}; further details on the pipeline are provided in Appendix~\ref{appendix:indexing}.

%\dataset{} is well-suited as a source in this method, as its diversity of context and basis in human consensus means th 

\subsection{\dataset{}-Enhanced Results}
To robustly test the utility of \dataset{} on new data, we curate an out-of-domain test dataset based on egocentric robotic assistant footage \cite{zhu2024siat}, selected as its context and embodiment are orthogonal to those seen in Ego4D. Actions and justifications are manually generated to be highly challenging, with baseline GPT-4o scoring 18.2\%.\footnote{11 samples were selected from 100 candidate samples, from which 11 datapoints were generated to maximize the diversity of actions and contexts represented. While this is a sufficient number for the purposes of this example, future work should target a wider range of embodiments.} Using retrieval across \dataset{}, we demonstrate improvement relative to the best non-RAG model and base GPT-4o on unseen in-domain tasks, obtaining an \dataset{} bench 9.4\% better than base GPT-4o, and 7.9\% better than randomized retrieval across \dataset{}, as shown in Table~\ref{tab:normthinker1}.

%By contrast, \normthinker{} exhibits a notable improvement in normative reasoning, as shown in Table \ref{tab:normthinker}. % , resulting in enhanced performance. 

%However, it still demonstrates a significant performance gap compared to human reasoning.
\begin{table}
\centering
\small
\begin{tabular}{l cccc}
\toprule
 \multirow{2}{*}{Model} & \multicolumn{3}{c}{\% Correct MCQ} & {Sens.} \\
\cmidrule(lr){2-5}
 & \multicolumn{1}{c}{Both} & \multicolumn{1}{c}{Act.} & \multicolumn{1}{c}{Jus.} &  \multicolumn{1}{c}{Act.} \\
\midrule
 {GPT-4o} & 1/11 & 5/11 & 2/11 & 3/11 \\
 {\quad + Best-5 Retrieval} & \textbf{5/11} & \textbf{7/11} & \textbf{5/11} & 3/11 \\ 
\midrule
\midrule
 {Human} & 8/11 & 8/11 & 8/11 & 9/11 \\ 
\bottomrule     
\end{tabular}
\caption{Results with \normthinker{} on egocentric robotics videos, n=11. }
\label{tab:normthinker}
\end{table}

\begin{table}
\centering
\small
\begin{tabular}{l cccc}
\toprule
 \multirow{2}{*}{Model} & \multicolumn{3}{c}{\% Correct MCQ} & {Sens.} \\
\cmidrule(lr){2-5}
 & \multicolumn{1}{c}{Both} & \multicolumn{1}{c}{Act.} & \multicolumn{1}{c}{Jus.} &  \multicolumn{1}{c}{Act.} \\
% Constant Choice &  25.3 & 25.3 & 25.3 & 40.5 \\
\midrule
{Gemini 1.5 Pro} & 45.2 & 51.8 & 47.7 & \textbf{64.0}\\
{GPT-4o} & 39.8 & 44.9 & 45.1 & 59.6 \\
{\quad + Random Retrieval} & 41.3 & 51.0 & 45.7 & 52.6 \\
{\quad + Best-5 Retrieval} & \textbf{49.2} & \textbf{54.5} &\textbf{52.6} & 56.2 \\ 
\midrule
{Human} & 92.4 & 92.4 & 92.4 & 85.1 \\ 
\bottomrule     
\end{tabular}
\caption{Results with \normthinker{} on held-out instances in \dataset{}.} 
\label{tab:normthinker1}
\end{table}

% One example \normthinker{} in action is illustrated in Figure~\ref{fig:normthinker_example}. In the case of unenriched inference on the example \dataset{} task, GPT-4o correctly selects the action but arrives at incorrect reasoning, concluding that one should play games with friends after finishing a meal. In contrast, with \normthinker{}, the top retrieved datapoint, which depicts a similar cleanup process following a game, provides an in-context example that contextualizes tidying up as a sign of respect as the most normative action in this class of context, enabling GPT-4o to rethink and select the correct justification: "help with cleanup."

% \begin{figure*}
%     \centering
%     \includegraphics[width=\linewidth]{img/normthinker_example.jpg}
%     \caption{An example from \normthinker{} illustrating how retrieved data points aid normative reasoning. The correct answers are underlined. Without the reference video and justification, GPT 4o selects the correct action but provides incorrect reasoning. With the retrieved data point—depicting a similar cleanup process, GPT 4o selects the correct justification: "help with cleanup."}

%     \label{fig:normthinker_example}
% \end{figure*}

\section{Related Work}
\label{sec:related_work}

% \citet{ziems-etal-2023-normbank} introduced NormBank

% \subsection{Social Navigation for Embodied Agent}
% Social navigation for embodied agents encompasses the integration of social norms and behaviors into the path-planning algorithms of robots and other forms of agents. Previous studies have pioneered the development of social force models \citep{helbing1995social, leal2011everybody}, deep reinforcement learning approaches \citep{chen2018socially, everett2018motion, sathyamoorthy2020frozone}, and inverse reinforcement learning \citep{Kretzschmar2016SociallyCM} to enable agents to navigate complex environments without causing discomfort to any human present. This task usually requires the agent to actively perceive the environment \citep{daza2021approach}, track and predict the movement of pedestrians \citep{leal2011everybody, Kruse2013HumanawareRN, sathyamoorthy2020frozone}， and even account for invisible human \citep{singamaneni2022watch}.

\subsection{Video Question Answering}
Video Question Answering has emerged as a widely adopted benchmark for VLMs, framing visual understanding as a question-answering task~\cite{lei2019tvqalocalizedcompositionalvideo, yu2019activitynetqadatasetunderstandingcomplex, xiao2021nextqanextphasequestionansweringexplaining,zhu2023excalibur}. Many benchmarks employ MCQ tasks to simplify evaluation by providing an aggregate accuracy metric~\cite{chandrasegaran2024hourvideo1hourvideolanguageunderstanding, chinchure2024blackswanabductivedefeasible}. For example, VCR~\cite{zellers2019recognition} introduces \textit{Adversarial Matching} to create challenging MCQs with minimal human intervention. HourVideo~\cite{chandrasegaran2024hourvideo1hourvideolanguageunderstanding} utilizes a five-stage pipeline to generate, refine, and filter diverse, high-quality MCQs. Similarly, EgoSchema~\cite{mangalam2023egoschemadiagnosticbenchmarklongform} leverages Ego4D~\cite{grauman2022ego4dworld3000hours} videos and implements several rounds of filtering and manual curation, to ensure that questions are both high-quality and sufficiently challenging~\cite{mangalam2023egoschemadiagnosticbenchmarklongform}.

\subsection{Social Commonsense and Norms}
Commonsense knowledge bases, such as ConceptNet~\citep{speer2017conceptnet} and ATOMIC~\citep{sap2019atomic}, provide AI systems with essential everyday information for tasks ranging from physical commonsense reasoning to explanation generation. NormBank~\citep{ziems-etal-2023-normbank} further enriches this landscape by offering situational contrast sets that support normative reasoning about unspoken social rules.
Complementing these resources, social intelligence benchmarks like the ToMi~\citep{le-etal-2019-revisiting} and FauxPas datasets~\citep{shapira-etal-2023-well}—along with simulation environments such as SOTOPIA~\citep{zhou2024sotopia, wang2024sotopiapiinteractivelearningsocially}—assess an agent’s ability to understand others' intentions and navigate complex social interactions.
Recent work has expanded these evaluations to embodied agents~\citep{kwon2024grounded, padmakumar2021teach} and diverse task scenarios ~\citep{wang-etal-2019-persuasion, Bakhtin2022HumanlevelPI}. Building on these insights, our work introduces a benchmark specifically designed to evaluate normative decision-making abilities.

\section{Conclusion}
%\hao{Future work? How should researchers improve norm reasoning?}
We introduce \dataset{}, a novel benchmark and dataset designed to rigorously evaluate the ability of VLMs to understand physical social norms (PSN) in egocentric embodiments. We demonstrate that, despite SOTA models' strong visual recognition and abstract reasoning capabilities, they remain inferior to humans in PSN understanding, primarily due to norm sensibility and prioritization errors. We demonstrate \dataset{}'s direct utility in augmenting normative understanding by testing a retrieval-based method, demonstrating improvements across out-of-domain and out-of-embodiment videos. Finally, we identify opportunities for future work in embodied norm understanding, suggesting post-training on large norm datasets as a promising direction for study.

% \pagebreak

\section*{Limitations}
While multiple rounds of filtering are applied to ensure diversity in ~\dataset{} video clips, all video clips in ~\dataset{} are exclusively from Ego4D, which may reflect inherent distribution biases within Ego4D. Expanding the benchmark to include a broader range of video sources, including exocentric videos, would improve the generalization of the benchmark.

\noindent Another limitation is that the current evaluation scheme treats videos as sequences of frames without incorporating audio information, which limits model performance on tasks that rely heavily on auditory cues. Integrating the audio modality in future work would provide a more comprehensive assessment of the normative reasoning abilities of vision-language models.

% , as well as more comprehensively represent the ranges of embodiments that agents might take.

% Another limitation is that the current evaluation treats videos as sequences of frames without incorporating audio information---this is a fundamental limitation of the current vision-language models; it was determined that testing on a wider, more representative range of models without access to audio was a worthwhile compromise to capture SOTA model performance. Integrating the audio modality, either natively or through an audio-to-text encoder, would provide a more comprehensive assessment of vision-language models, particularly for tasks and norms that rely on auditory cues.

% \yicheng{ego vs exo centric videos are discussed in paragraph 1 I think}
% \hao{Looks good, just add some backlink references to the section where we did the filtering}
\noindent Finally, though the generation and filtering pipeline (\S\ref{sec:benchmark_generation_pipeline}) is robust in generating high-difficulty and high-quality \dataset{} tasks, we find that Ego4D contains many action annotation errors that could lead to the generation of ambiguous or incorrect MCQs. We thus carefully conduct additional manual multi-stage filtering processes and human validation to remove or rectify low-quality samples from ~\dataset{} to mitigate the impact of this issue.
\section*{Ethics Statement}
\label{sec:ethics}
\paragraph{Ethical Assumptions.} 
We emphasize that \dataset{} is designed as a descriptive benchmark rather than a prescriptive one --- 
the dataset is intended to evaluate the ability of VLMs to understand physical social norms in egocentric videos, rather than to dictate what these norms should be or how they should be enforced.
We thus acknowledge that the norms depicted in the dataset may not be universally applicable or appropriate in all contexts and that the interpretation of these norms may vary across cultures, communities, time periods, and individuals.

\paragraph{Bias and Fairness.}
Despite our best efforts to create a diverse and representative dataset, 
we acknowledge that \dataset{} may contain biases that reflect the perspectives and experiences of the dataset creators and annotators.
Consequently, the norms and justifications depicted in the dataset may be influenced by the cultural, social, and demographic characteristics of the individuals who contributed to the dataset.
While all of our annotators are from the United States, norms often differ in different cultures \citep{rao2024normad, shi2024culturebank}.
To address these concerns, 
we recommend that researchers using \dataset{} for training or evaluation critically assess potential biases and ensure they align with the intended application context.

\paragraph{Human Subjects and Privacy.}
\dataset{} is constructed from Ego4D videos, 
which are publicly available and do not contain personally identifiable information.
The Ego4D dataset is released under a non-exclusive, non-transferable license that permits its use for academic research, as outlined in the license agreement. Our work complies with the terms of this license, using the Ego4D data solely for research purposes. Our annotation process was conducted with proper informed consent,
ensuring annotators are fully aware of the task, its purpose, and how their contribution would be used.
Annotators were compensated fairly for their time and effort (details in Appendix~\ref{sec:HumanValidationProcess}).
The data used in this work does not include personally identifiable information.
No sensitive information about the annotators or individuals appearing in the video data was collected or used in the study.
% This work was reviewed and deemed exempt by the Institutional Review Board at the institution where the work was conducted. %\mohammad{I'm not sure about this part.}
Notably,
this work was thoroughly reviewed and approved by the Institutional Review Board (IRB) at Stanford University (IRB-77185).

\paragraph{Risks in Deployment.}
The deployment of AI systems trained on \dataset{} may pose risks if these systems are used to make decisions that impact individuals' safety, well-being, or rights.
To mitigate these risks, 
we stress that \dataset{} should not be used for prescriptive advice or to make decisions with ethical, or safety implications without extensive human oversight. 
By using \dataset{},
researchers should be aware of the limitations of the dataset and the potential risks associated with deploying systems trained on it.
\section*{Acknowledgments}
This research was supported in part by Other Transaction award HR00112490375 from the U.S. Defense Advanced Research Projects Agency (DARPA) Friction for Accountability in Conversational Transactions (FACT) program. We thank Google Cloud Platform and Modal Platform for their credits. We also thank Yonatan Bisk, Dorsa Sadigh, and members of the Stanford SALT lab for their feedback and input. The authors thank Leena Mathur and Su Li for their help in collecting out-of-domain robotics videos.

\bibliography{custom}

\begin{thebibliography}{74}
\providecommand{\natexlab}[1]{#1}

\bibitem[{Altman(1975)}]{altman1975environment}
Irwin Altman. 1975.
\newblock The environment and social behavior: privacy, personal space, territory, and crowding.

\bibitem[{Anthropic(2024)}]{anthropic2024claude}
Anthropic. 2024.
\newblock \href {https://www-cdn.anthropic.com/de8ba9b01c9ab7cbabf5c33b80b7bbc618857627/Model_Card_Claude_3.pdf} {The claude 3 model family: Opus, sonnet, haiku}.

\bibitem[{Asimov(1985)}]{asimov1985caves}
Isaac Asimov. 1985.
\newblock \emph{The caves of steel}.
\newblock Random House Publishing Group.

\bibitem[{Bakhtin et~al.(2022)Bakhtin, Brown, Dinan, Farina, Flaherty, Fried, Goff, Gray, Hu, Jacob, Komeili, Konath, Kwon, Lerer, Lewis, Miller, Mitts, Renduchintala, Roller, Rowe, Shi, Spisak, Wei, Wu, Zhang, and Zijlstra}]{Bakhtin2022HumanlevelPI}
Anton Bakhtin, Noam Brown, Emily Dinan, Gabriele Farina, Colin Flaherty, Daniel Fried, Andrew Goff, Jonathan Gray, Hengyuan Hu, Athul~Paul Jacob, Mojtaba Komeili, Karthik Konath, Minae Kwon, Adam Lerer, Mike Lewis, Alexander~H. Miller, Sandra Mitts, Adithya Renduchintala, Stephen Roller, Dirk Rowe, Weiyan Shi, Joe Spisak, Alexander Wei, David~J. Wu, Hugh Zhang, and Markus Zijlstra. 2022.
\newblock \href {https://api.semanticscholar.org/CorpusID:253759631} {Human-level play in the game of diplomacy by combining language models with strategic reasoning}.
\newblock \emph{Science}, 378:1067 -- 1074.

\bibitem[{Chambers(2016)}]{chambers2016closed}
Becky Chambers. 2016.
\newblock \emph{A Closed and Common Orbit}.
\newblock Hodder \& Stoughton.

\bibitem[{Chandrasegaran et~al.(2024)Chandrasegaran, Gupta, Hadzic, Kota, He, Eyzaguirre, Durante, Li, Wu, and Li}]{chandrasegaran2024hourvideo1hourvideolanguageunderstanding}
Keshigeyan Chandrasegaran, Agrim Gupta, Lea~M. Hadzic, Taran Kota, Jimming He, Cristobal Eyzaguirre, Zane Durante, Manling Li, Jiajun Wu, and Fei-Fei Li. 2024.
\newblock Hourvideo: 1-hour video-language understanding.
\newblock In \emph{Advances in Neural Information Processing Systems}, volume~37.

\bibitem[{Chang et~al.(2024)Chang, Chhablani, Clegg, Cote, Desai, Hlavac, Karashchuk, Krantz, Mottaghi, Parashar et~al.}]{chang2024partnr}
Matthew Chang, Gunjan Chhablani, Alexander Clegg, Mikael~Dallaire Cote, Ruta Desai, Michal Hlavac, Vladimir Karashchuk, Jacob Krantz, Roozbeh Mottaghi, Priyam Parashar, et~al. 2024.
\newblock Partnr: A benchmark for planning and reasoning in embodied multi-agent tasks.
\newblock \emph{arXiv preprint arXiv:2411.00081}.

\bibitem[{Chen et~al.(2024{\natexlab{a}})Chen, Xu, Kirmani, Ichter, Sadigh, Guibas, and Xia}]{chen2024spatialvlm}
Boyuan Chen, Zhuo Xu, Sean Kirmani, Brian Ichter, Dorsa Sadigh, Leonidas Guibas, and Fei Xia. 2024{\natexlab{a}}.
\newblock \href {https://openaccess.thecvf.com/content/CVPR2024/papers/Chen_SpatialVLM_Endowing_Vision-Language_Models_with_Spatial_Reasoning_Capabilities_CVPR_2024_paper.pdf} {Spatialvlm: Endowing vision-language models with spatial reasoning capabilities}.
\newblock In \emph{Proceedings of the IEEE/CVF Conference on Computer Vision and Pattern Recognition (CVPR)}, pages 14455--14465. IEEE.

\bibitem[{Chen et~al.(2024{\natexlab{b}})Chen, Wang, Cao, Liu, Gao, Cui, Zhu, Ye, Tian, Liu, Gu, Wang, Li, Ren, Chen, Luo, Wang, Jiang, Wang, He, Shi, Zhang, Lv, Wang, Shao, Chu, Tu, He, Wu, Deng, Ge, Chen, Zhang, Wang, Dou, Lu, Zhu, Lu, Lin, Qiao, Dai, and Wang}]{chen2024expanding}
Zhe Chen, Weiyun Wang, Yue Cao, Yangzhou Liu, Zhangwei Gao, Erfei Cui, Jinguo Zhu, Shenglong Ye, Hao Tian, Zhaoyang Liu, Lixin Gu, Xuehui Wang, Qingyun Li, Yimin Ren, Zixuan Chen, Jiapeng Luo, Jiahao Wang, Tan Jiang, Bo~Wang, Conghui He, Botian Shi, Xingcheng Zhang, Han Lv, Yi~Wang, Wenqi Shao, Pei Chu, Zhongying Tu, Tong He, Zhiyong Wu, Huipeng Deng, Jiaye Ge, Kai Chen, Kaipeng Zhang, Limin Wang, Min Dou, Lewei Lu, Xizhou Zhu, Tong Lu, Dahua Lin, Yu~Qiao, Jifeng Dai, and Wenhai Wang. 2024{\natexlab{b}}.
\newblock \href {https://arxiv.org/abs/2412.05271} {Expanding performance boundaries of open-source multimodal models with model, data, and test-time scaling}.
\newblock \emph{Preprint}, arXiv:2412.05271.

\bibitem[{Chiang(2010)}]{chiang2010lifecycle}
Ted Chiang. 2010.
\newblock \emph{The lifecycle of software objects}.
\newblock Subterranean Press Burton.

\bibitem[{Chinchure et~al.(2024)Chinchure, Ravi, Ng, Shwartz, Li, and Sigal}]{chinchure2024blackswanabductivedefeasible}
Aditya Chinchure, Sahithya Ravi, Raymond Ng, Vered Shwartz, Boyang Li, and Leonid Sigal. 2024.
\newblock \href {https://arxiv.org/abs/2412.05725} {Black swan: Abductive and defeasible video reasoning in unpredictable events}.
\newblock \emph{Preprint}, arXiv:2412.05725.

\bibitem[{Chollet et~al.(2024)Chollet, Knoop, Kamradt, and Landers}]{chollet2024arc}
Francois Chollet, Mike Knoop, Gregory Kamradt, and Bryan Landers. 2024.
\newblock Arc prize 2024: Technical report.
\newblock \emph{arXiv preprint arXiv:2412.04604}.

\bibitem[{Chow et~al.(2025)Chow, Mao, Li, Seita, Guizilini, and Wang}]{chow2025physbench}
Wei Chow, Jiageng Mao, Boyi Li, Daniel Seita, Vitor Guizilini, and Yue Wang. 2025.
\newblock \href {https://arxiv.org/abs/2501.16411} {Physbench: Benchmarking and enhancing vision-language models for physical world understanding}.
\newblock In \emph{Proceedings of the International Conference on Learning Representations (ICLR)}.

\bibitem[{Chudek and Henrich(2011)}]{chudek2011culture}
Maciej Chudek and Joseph Henrich. 2011.
\newblock Culture--gene coevolution, norm-psychology and the emergence of human prosociality.
\newblock \emph{Trends in cognitive sciences}, 15(5):218--226.

\bibitem[{Chung and Rimal(2016)}]{chung2016social}
Adrienne Chung~Adrienne Chung and Rajiv N Rimal Rajiv~N Rimal. 2016.
\newblock Social norms: A review.
\newblock \emph{Review of Communication Research}, 4:01--28.

\bibitem[{Fehr and Fischbacher(2004)}]{fehr2004social}
Ernst Fehr and Urs Fischbacher. 2004.
\newblock Social norms and human cooperation.
\newblock \emph{Trends in cognitive sciences}, 8(4):185--190.

\bibitem[{Francis et~al.(2024)Francis, P{\'e}rez-d'Arpino, Li, Xia, Alahi, Bera, Biswas, Biswas, Chandra, Lewis~Chiang, Everett, Ha, Hart, How, Karnan, Lee, Manso, Mirksy, Pirk, Stone, Taylor, Trautman, Tsoi, V{\'a}zquez, Xiao, Xu, Yokoyama, Toshev, Mart{\'i}n-Mart{\'i}n, Alami, and Singamaneni}]{francis2023principles}
Anthony Francis, Claudia P{\'e}rez-d'Arpino, Chengshu Li, Fei Xia, Alexandre Alahi, Aniket Bera, Abhijat Biswas, Joydeep Biswas, Rohan Chandra, Hao-Tien Lewis~Chiang, Michael Everett, Sehoon Ha, Justin Hart, Jonathan~P How, Haresh Karnan, Tsang-Wei~Edward Lee, Luis~J Manso, Reuth Mirksy, S{\"o}ren Pirk, Peter Stone, Ada~V Taylor, Peter Trautman, Nathan Tsoi, Marynel V{\'a}zquez, Xuesu Xiao, Peng Xu, Naoki Yokoyama, Alexander Toshev, Roberto Mart{\'i}n-Mart{\'i}n, Rachid Alami, and Phani-Teja Singamaneni. 2024.
\newblock \href {https://doi.org/10.1145/3700599} {{Principles and Guidelines for Evaluating Social Robot Navigation Algorithms}}.
\newblock \emph{{ACM Transactions on Human-Robot Interaction}}.

\bibitem[{Francis et~al.(2023)Francis, Pérez-D'Arpino, Li, Xia, Alahi, Alami, Bera, Biswas, Biswas, Chandra, Chiang, Everett, Ha, Hart, How, Karnan, Lee, Manso, Mirksy, Pirk, Singamaneni, Stone, Taylor, Trautman, Tsoi, Vázquez, Xiao, Xu, Yokoyama, Toshev, and Martín-Martín}]{francis2023principlesguidelinesevaluatingsocial}
Anthony Francis, Claudia Pérez-D'Arpino, Chengshu Li, Fei Xia, Alexandre Alahi, Rachid Alami, Aniket Bera, Abhijat Biswas, Joydeep Biswas, Rohan Chandra, Hao-Tien~Lewis Chiang, Michael Everett, Sehoon Ha, Justin Hart, Jonathan~P. How, Haresh Karnan, Tsang-Wei~Edward Lee, Luis~J. Manso, Reuth Mirksy, Sören Pirk, Phani~Teja Singamaneni, Peter Stone, Ada~V. Taylor, Peter Trautman, Nathan Tsoi, Marynel Vázquez, Xuesu Xiao, Peng Xu, Naoki Yokoyama, Alexander Toshev, and Roberto Martín-Martín. 2023.
\newblock \href {https://arxiv.org/abs/2306.16740} {Principles and guidelines for evaluating social robot navigation algorithms}.
\newblock \emph{Preprint}, arXiv:2306.16740.

\bibitem[{Gibbs(1965)}]{gibbs1965norms}
Jack~P Gibbs. 1965.
\newblock Norms: The problem of definition and classification.
\newblock \emph{American Journal of Sociology}, 70(5):586--594.

\bibitem[{Goyal et~al.(2017)Goyal, Khot, Summers-Stay, Batra, and Parikh}]{Goyal_2017_CVPR}
Yash Goyal, Tejas Khot, Douglas Summers-Stay, Dhruv Batra, and Devi Parikh. 2017.
\newblock Making the v in vqa matter: Elevating the role of image understanding in visual question answering.
\newblock In \emph{Proceedings of the IEEE Conference on Computer Vision and Pattern Recognition (CVPR)}.

\bibitem[{Grauman et~al.(2022)Grauman, Westbury, Byrne, Chavis, Furnari, Girdhar, Hamburger, Jiang, Liu, Liu, Martin, Nagarajan, Radosavovic, Ramakrishnan, Ryan, Sharma, Wray, Xu, Xu, Zhao, Bansal, Batra, Cartillier, Crane, Do, Doulaty, Erapalli, Feichtenhofer, Fragomeni, Fu, Gebreselasie, Gonzalez, Hillis, Huang, Huang, Jia, Khoo, Kolar, Kottur, Kumar, Landini, Li, Li, Li, Mangalam, Modhugu, Munro, Murrell, Nishiyasu, Price, Puentes, Ramazanova, Sari, Somasundaram, Southerland, Sugano, Tao, Vo, Wang, Wu, Yagi, Zhao, Zhu, Arbelaez, Crandall, Damen, Farinella, Fuegen, Ghanem, Ithapu, Jawahar, Joo, Kitani, Li, Newcombe, Oliva, Park, Rehg, Sato, Shi, Shou, Torralba, Torresani, Yan, and Malik}]{grauman2022ego4dworld3000hours}
Kristen Grauman, Andrew Westbury, Eugene Byrne, Zachary Chavis, Antonino Furnari, Rohit Girdhar, Jackson Hamburger, Hao Jiang, Miao Liu, Xingyu Liu, Miguel Martin, Tushar Nagarajan, Ilija Radosavovic, Santhosh~Kumar Ramakrishnan, Fiona Ryan, Jayant Sharma, Michael Wray, Mengmeng Xu, Eric~Zhongcong Xu, Chen Zhao, Siddhant Bansal, Dhruv Batra, Vincent Cartillier, Sean Crane, Tien Do, Morrie Doulaty, Akshay Erapalli, Christoph Feichtenhofer, Adriano Fragomeni, Qichen Fu, Abrham Gebreselasie, Cristina Gonzalez, James Hillis, Xuhua Huang, Yifei Huang, Wenqi Jia, Weslie Khoo, Jachym Kolar, Satwik Kottur, Anurag Kumar, Federico Landini, Chao Li, Yanghao Li, Zhenqiang Li, Karttikeya Mangalam, Raghava Modhugu, Jonathan Munro, Tullie Murrell, Takumi Nishiyasu, Will Price, Paola~Ruiz Puentes, Merey Ramazanova, Leda Sari, Kiran Somasundaram, Audrey Southerland, Yusuke Sugano, Ruijie Tao, Minh Vo, Yuchen Wang, Xindi Wu, Takuma Yagi, Ziwei Zhao, Yunyi Zhu, Pablo Arbelaez, David Crandall, Dima Damen, Giovanni~Maria
  Farinella, Christian Fuegen, Bernard Ghanem, Vamsi~Krishna Ithapu, C.~V. Jawahar, Hanbyul Joo, Kris Kitani, Haizhou Li, Richard Newcombe, Aude Oliva, Hyun~Soo Park, James~M. Rehg, Yoichi Sato, Jianbo Shi, Mike~Zheng Shou, Antonio Torralba, Lorenzo Torresani, Mingfei Yan, and Jitendra Malik. 2022.
\newblock \href {https://arxiv.org/abs/2110.07058} {Ego4d: Around the world in 3,000 hours of egocentric video}.
\newblock \emph{Preprint}, arXiv:2110.07058.

\bibitem[{Guo et~al.(2025)Guo, Yang, Zhang, Song, Zhang, Xu, Zhu, Ma, Wang, Bi et~al.}]{guo2025deepseek}
Daya Guo, Dejian Yang, Haowei Zhang, Junxiao Song, Ruoyu Zhang, Runxin Xu, Qihao Zhu, Shirong Ma, Peiyi Wang, Xiao Bi, et~al. 2025.
\newblock Deepseek-r1: Incentivizing reasoning capability in llms via reinforcement learning.
\newblock \emph{arXiv preprint arXiv:2501.12948}.

\bibitem[{He et~al.(2024)He, Zhu, Ye, Liu, Zhou, and Yu}]{he2024emerged}
Feng He, Tianqing Zhu, Dayong Ye, Bo~Liu, Wanlei Zhou, and Philip~S Yu. 2024.
\newblock The emerged security and privacy of llm agent: A survey with case studies.
\newblock \emph{arXiv preprint arXiv:2407.19354}.

\bibitem[{Hechter and Opp(2001)}]{hechter2001social}
Michael Hechter and Karl-Dieter Opp. 2001.
\newblock Social norms.

\bibitem[{Hollander and Wu(2011)}]{hollander2011current}
Christopher~D Hollander and Annie~S Wu. 2011.
\newblock The current state of normative agent-based systems.
\newblock \emph{Journal of Artificial Societies and Social Simulation}, 14(2):6.

\bibitem[{Huang et~al.(2022)Huang, Knierim, Chiossi, Chuang, and Welsch}]{huang2022proxemics}
Ann Huang, Pascal Knierim, Francesco Chiossi, Lewis~L Chuang, and Robin Welsch. 2022.
\newblock Proxemics for human-agent interaction in augmented reality.
\newblock In \emph{Proceedings of the 2022 CHI conference on human factors in computing systems}, pages 1--13.

\bibitem[{Hurst et~al.(2024)Hurst, Lerer, Goucher, Perelman, Ramesh, Clark, Ostrow, Welihinda, Hayes, Radford et~al.}]{hurst2024gpt}
Aaron Hurst, Adam Lerer, Adam~P Goucher, Adam Perelman, Aditya Ramesh, Aidan Clark, AJ~Ostrow, Akila Welihinda, Alan Hayes, Alec Radford, et~al. 2024.
\newblock Gpt-4o system card.
\newblock \emph{arXiv preprint arXiv:2410.21276}.

\bibitem[{Jaech et~al.(2024)Jaech, Kalai, Lerer, Richardson, El-Kishky, Low, Helyar, Madry, Beutel, Carney et~al.}]{jaech2024openai}
Aaron Jaech, Adam Kalai, Adam Lerer, Adam Richardson, Ahmed El-Kishky, Aiden Low, Alec Helyar, Aleksander Madry, Alex Beutel, Alex Carney, et~al. 2024.
\newblock Openai o1 system card.
\newblock \emph{arXiv preprint arXiv:2412.16720}.

\bibitem[{John(2006)}]{john2006android}
Scalzi John. 2006.
\newblock \emph{The android's dream}.
\newblock A Tom Doherty Associates Books, New York.

\bibitem[{K{\"o}ster and Hepach(2024)}]{koster2024preverbal}
Moritz K{\"o}ster and Robert Hepach. 2024.
\newblock Preverbal infants’ understanding of social norms.
\newblock \emph{Scientific Reports}, 14(1):2983.

\bibitem[{Kwon et~al.(2024)Kwon, Hu, Myers, Karamcheti, Dragan, and Sadigh}]{kwon2024grounded}
Minae Kwon, Hengyuan Hu, Vivek Myers, Siddharth Karamcheti, Anca Dragan, and Dorsa Sadigh. 2024.
\newblock \href {https://arxiv.org/abs/2306.08651} {Toward grounded commonsense reasoning}.
\newblock \emph{Preprint}, arXiv:2306.08651.

\bibitem[{Lasota et~al.(2017)Lasota, Fong, Shah et~al.}]{lasota2017survey}
Przemyslaw~A Lasota, Terrence Fong, Julie~A Shah, et~al. 2017.
\newblock A survey of methods for safe human-robot interaction.
\newblock \emph{Foundations and Trends{\textregistered} in Robotics}, 5(4):261--349.

\bibitem[{Le et~al.(2019)Le, Boureau, and Nickel}]{le-etal-2019-revisiting}
Matthew Le, Y-Lan Boureau, and Maximilian Nickel. 2019.
\newblock \href {https://doi.org/10.18653/v1/D19-1598} {Revisiting the evaluation of theory of mind through question answering}.
\newblock In \emph{Proceedings of the 2019 Conference on Empirical Methods in Natural Language Processing and the 9th International Joint Conference on Natural Language Processing (EMNLP-IJCNLP)}, pages 5872--5877, Hong Kong, China. Association for Computational Linguistics.

\bibitem[{Lei et~al.(2018)Lei, Yu, Bansal, and Berg}]{lei2019tvqalocalizedcompositionalvideo}
Jie Lei, Licheng Yu, Mohit Bansal, and Tamara Berg. 2018.
\newblock \href {https://doi.org/10.18653/v1/D18-1167} {{TVQA}: Localized, compositional video question answering}.
\newblock In \emph{Proceedings of the 2018 Conference on Empirical Methods in Natural Language Processing}, pages 1369--1379, Brussels, Belgium. Association for Computational Linguistics.

\bibitem[{Lewis et~al.(2020)Lewis, Perez, Piktus, Petroni, Rockt{\"a}schel, Ruder, Weihs, and Kiela}]{lewis2020retrieval}
Patrick Lewis, Ethan Perez, Aleksandra Piktus, Fabio Petroni, Tim Rockt{\"a}schel, Sebastian Ruder, Luca Weihs, and Douwe Kiela. 2020.
\newblock \href {https://arxiv.org/abs/2005.11401} {Retrieval-augmented generation for knowledge-intensive nlp tasks}.
\newblock \emph{arXiv preprint arXiv:2005.11401}.

\bibitem[{Li et~al.(2024{\natexlab{a}})Li, Gan, Yang, Yang, Li, Wang, Gao et~al.}]{li2024multimodal}
Chunyuan Li, Zhe Gan, Zhengyuan Yang, Jianwei Yang, Linjie Li, Lijuan Wang, Jianfeng Gao, et~al. 2024{\natexlab{a}}.
\newblock Multimodal foundation models: From specialists to general-purpose assistants.
\newblock \emph{Foundations and Trends{\textregistered} in Computer Graphics and Vision}, 16(1-2):1--214.

\bibitem[{Li et~al.(2024{\natexlab{b}})Li, Zhao, Wang, Wang, Zhou, Srivastava, Gokmen, Lee, Li, Zhang et~al.}]{liembodied}
Manling Li, Shiyu Zhao, Qineng Wang, Kangrui Wang, Yu~Zhou, Sanjana Srivastava, Cem Gokmen, Tony Lee, Li~Erran Li, Ruohan Zhang, et~al. 2024{\natexlab{b}}.
\newblock Embodied agent interface: Benchmarking llms for embodied decision making.
\newblock In \emph{The Thirty-eight Conference on Neural Information Processing Systems Datasets and Benchmarks Track}.

\bibitem[{Lutz and Tam{\'o}-Larrieux(2020)}]{lutz2020robot}
Christoph Lutz and Aurelia Tam{\'o}-Larrieux. 2020.
\newblock \href {https://doi.org/10.30658/hmc.1.6} {The robot privacy paradox: Understanding how privacy concerns shape intentions to use social robots}.
\newblock \emph{Human-Machine Communication}, 1:87--104.

\bibitem[{Mangalam et~al.(2023)Mangalam, Akshulakov, and Malik}]{mangalam2023egoschemadiagnosticbenchmarklongform}
Karttikeya Mangalam, Raiymbek Akshulakov, and Jitendra Malik. 2023.
\newblock \href {https://proceedings.neurips.cc/paper_files/paper/2023/file/90ce332aff156b910b002ce4e6880dec-Paper-Datasets_and_Benchmarks.pdf} {Egoschema: A diagnostic benchmark for very long-form video language understanding}.
\newblock In \emph{Advances in Neural Information Processing Systems}, volume~36, pages 46212--46244. Curran Associates, Inc.

\bibitem[{Mavrogiannis et~al.(2023)Mavrogiannis, Baldini, Wang, Zhao, Trautman, Steinfeld, and Oh}]{mavrogiannis2023core}
Christoforos Mavrogiannis, Francesca Baldini, Allan Wang, Dapeng Zhao, Pete Trautman, Aaron Steinfeld, and Jean Oh. 2023.
\newblock Core challenges of social robot navigation: A survey.
\newblock \emph{ACM Transactions on Human-Robot Interaction}, 12(3):1--39.

\bibitem[{Mills and K{\'a}d{\'a}r(2011)}]{mills2011politeness}
Sara Mills and D{\'a}niel~Z K{\'a}d{\'a}r. 2011.
\newblock Politeness and culture.
\newblock \emph{Politeness in East Asia}, pages 21--44.

\bibitem[{Morsky and Ak{\c{c}}ay(2019)}]{morsky2019evolution}
Bryce Morsky and Erol Ak{\c{c}}ay. 2019.
\newblock Evolution of social norms and correlated equilibria.
\newblock \emph{Proceedings of the National Academy of Sciences}, 116(18):8834--8839.

\bibitem[{Mukherjee et~al.(2007)Mukherjee, Sen, and Airiau}]{mukherjee2007emergence}
Partha Mukherjee, Sandip Sen, and Stephane Airiau. 2007.
\newblock Emergence of norms with biased interactions in heterogeneous agent societies.
\newblock In \emph{2007 IEEE/WIC/ACM International Conferences on Web Intelligence and Intelligent Agent Technology-Workshops}, pages 512--515. IEEE.

\bibitem[{Neggers et~al.(2022)Neggers, Cuijpers, Ruijten, and IJsselsteijn}]{neggers2022determining}
Margot~ME Neggers, Raymond~H Cuijpers, Peter~AM Ruijten, and Wijnand~A IJsselsteijn. 2022.
\newblock Determining shape and size of personal space of a human when passed by a robot.
\newblock \emph{International Journal of Social Robotics}, 14(2):561--572.

\bibitem[{OpenAI(2024)}]{o3mini}
OpenAI. 2024.
\newblock \href {https://openai.com/index/o3-mini-system-card/} {[link]}.

\bibitem[{Ostrom(2000)}]{ostrom2000collective}
Elinor Ostrom. 2000.
\newblock Collective action and the evolution of social norms.
\newblock \emph{Journal of economic perspectives}, 14(3):137--158.

\bibitem[{Padmakumar et~al.(2021)Padmakumar, Thomason, Shrivastava, Lange, Narayan-Chen, Gella, Piramuthu, Tur, and Hakkani-Tur}]{padmakumar2021teach}
Aishwarya Padmakumar, Jesse Thomason, Ayush Shrivastava, Patrick Lange, Anjali Narayan-Chen, Spandana Gella, Robinson Piramuthu, Gokhan Tur, and Dilek Hakkani-Tur. 2021.
\newblock \href {https://arxiv.org/abs/2110.00534} {Teach: Task-driven embodied agents that chat}.
\newblock \emph{Preprint}, arXiv:2110.00534.

\bibitem[{Paternotte and Grose(2013)}]{paternotte2013social}
C{\'e}dric Paternotte and Jonathan Grose. 2013.
\newblock Social norms and game theory: Harmony or discord?
\newblock \emph{The British journal for the philosophy of science}.

\bibitem[{Rao et~al.(2024)Rao, Yerukola, Shah, Reinecke, and Sap}]{rao2024normad}
Abhinav Rao, Akhila Yerukola, Vishwa Shah, Katharina Reinecke, and Maarten Sap. 2024.
\newblock \href {https://arxiv.org/abs/2404.12464} {Normad: A framework for measuring the cultural adaptability of large language models}.
\newblock \emph{Preprint}, arXiv:2404.12464.

\bibitem[{Russell and Ward(1982)}]{russell1982environmental}
James~A Russell and Lawrence~M Ward. 1982.
\newblock Environmental psychology.
\newblock \emph{Annual review of psychology}.

\bibitem[{Sap et~al.(2019)Sap, Le~Bras, Allaway, Bhagavatula, Lourie, Rashkin, Roof, Smith, and Choi}]{sap2019atomic}
Maarten Sap, Ronan Le~Bras, Emily Allaway, Chandra Bhagavatula, Nicholas Lourie, Hannah Rashkin, Brendan Roof, Noah~A Smith, and Yejin Choi. 2019.
\newblock Atomic: An atlas of machine commonsense for if-then reasoning.
\newblock In \emph{Proceedings of the AAAI conference on artificial intelligence}, volume~33, pages 3027--3035.

\bibitem[{Schmidt et~al.(2016)Schmidt, Butler, Heinz, and Tomasello}]{schmidt2016young}
Marco~FH Schmidt, Lucas~P Butler, Julia Heinz, and Michael Tomasello. 2016.
\newblock Young children see a single action and infer a social norm: Promiscuous normativity in 3-year-olds.
\newblock \emph{Psychological Science}, 27(10):1360--1370.

\bibitem[{Shao et~al.(2024)Shao, Li, Shi, Liu, and Yang}]{shao2024privacylensevaluatingprivacynorm}
Yijia Shao, Tianshi Li, Weiyan Shi, Yanchen Liu, and Diyi Yang. 2024.
\newblock \href {https://arxiv.org/abs/2409.00138} {Privacylens: Evaluating privacy norm awareness of language models in action}.
\newblock \emph{Preprint}, arXiv:2409.00138.

\bibitem[{Shapira et~al.(2023)Shapira, Zwirn, and Goldberg}]{shapira-etal-2023-well}
Natalie Shapira, Guy Zwirn, and Yoav Goldberg. 2023.
\newblock \href {https://doi.org/10.18653/v1/2023.findings-acl.663} {How well do large language models perform on faux pas tests?}
\newblock In \emph{Findings of the Association for Computational Linguistics: ACL 2023}, pages 10438--10451, Toronto, Canada. Association for Computational Linguistics.

\bibitem[{Shi et~al.(2024)Shi, Li, Zhang, Ziems, Horesh, de~Paula, Yang et~al.}]{shi2024culturebank}
Weiyan Shi, Ryan Li, Yutong Zhang, Caleb Ziems, Raya Horesh, Rog{\'e}rio~Abreu de~Paula, Diyi Yang, et~al. 2024.
\newblock Culturebank: An online community-driven knowledge base towards culturally aware language technologies.
\newblock \emph{arXiv preprint arXiv:2404.15238}.

\bibitem[{Snell et~al.(2024)Snell, Lee, Xu, and Kumar}]{snell2024inferencescaling}
Charlie Snell, Jaehoon Lee, Kelvin Xu, and Aviral Kumar. 2024.
\newblock \href {https://arxiv.org/abs/2408.03314} {Scaling llm test-time compute optimally can be more effective than scaling model parameters}.
\newblock \emph{Preprint}, arXiv:2408.03314.

\bibitem[{Speer et~al.(2017)Speer, Chin, and Havasi}]{speer2017conceptnet}
Robyn Speer, Joshua Chin, and Catherine Havasi. 2017.
\newblock Conceptnet 5.5: An open multilingual graph of general knowledge.
\newblock In \emph{Proceedings of the AAAI conference on artificial intelligence}, volume~31.

\bibitem[{Sunstein(1996)}]{sunstein1996social}
Cass~R Sunstein. 1996.
\newblock Social norms and social roles.
\newblock \emph{Colum. L. Rev.}, 96:903.

\bibitem[{Team et~al.(2024)Team, Georgiev, Lei, Burnell, Bai, Gulati, Tanzer, Vincent, Pan, Wang et~al.}]{team2024gemini}
Gemini Team, Petko Georgiev, Ving~Ian Lei, Ryan Burnell, Libin Bai, Anmol Gulati, Garrett Tanzer, Damien Vincent, Zhufeng Pan, Shibo Wang, et~al. 2024.
\newblock Gemini 1.5: Unlocking multimodal understanding across millions of tokens of context.
\newblock \emph{arXiv preprint arXiv:2403.05530}.

\bibitem[{Team(2025)}]{Qwen2.5-VL}
Qwen Team. 2025.
\newblock \href {https://qwenlm.github.io/blog/qwen2.5-vl/} {Qwen2.5-vl}.

\bibitem[{Van~Lange et~al.(2013)Van~Lange, Joireman, Parks, and Van~Dijk}]{van2013psychology}
Paul~AM Van~Lange, Jeff Joireman, Craig~D Parks, and Eric Van~Dijk. 2013.
\newblock The psychology of social dilemmas: A review.
\newblock \emph{Organizational Behavior and Human Decision Processes}, 120(2):125--141.

\bibitem[{Wang et~al.(2024)Wang, Yu, Zhang, Qi, Sap, Bisk, Neubig, and Zhu}]{wang2024sotopiapiinteractivelearningsocially}
Ruiyi Wang, Haofei Yu, Wenxin Zhang, Zhengyang Qi, Maarten Sap, Yonatan Bisk, Graham Neubig, and Hao Zhu. 2024.
\newblock \href {https://doi.org/10.18653/v1/2024.acl-long.698} {{SOTOPIA}-{\ensuremath{\pi}}: Interactive learning of socially intelligent language agents}.
\newblock In \emph{Proceedings of the 62nd Annual Meeting of the Association for Computational Linguistics (Volume 1: Long Papers)}, pages 12912--12940, Bangkok, Thailand. Association for Computational Linguistics.

\bibitem[{Wang et~al.(2019)Wang, Shi, Kim, Oh, Yang, Zhang, and Yu}]{wang-etal-2019-persuasion}
Xuewei Wang, Weiyan Shi, Richard Kim, Yoojung Oh, Sijia Yang, Jingwen Zhang, and Zhou Yu. 2019.
\newblock \href {https://doi.org/10.18653/v1/P19-1566} {Persuasion for good: Towards a personalized persuasive dialogue system for social good}.
\newblock In \emph{Proceedings of the 57th Annual Meeting of the Association for Computational Linguistics}, pages 5635--5649, Florence, Italy. Association for Computational Linguistics.

\bibitem[{Wei et~al.(2022)Wei, Wang, Schuurmans, Bosma, Xia, Chi, Le, Zhou et~al.}]{wei2022chain}
Jason Wei, Xuezhi Wang, Dale Schuurmans, Maarten Bosma, Fei Xia, Ed~Chi, Quoc~V Le, Denny Zhou, et~al. 2022.
\newblock Chain-of-thought prompting elicits reasoning in large language models.
\newblock \emph{Advances in neural information processing systems}, 35:24824--24837.

\bibitem[{Wu et~al.(2024)Wu, Sun, Li, Welleck, and Yang}]{wu2024inferencescaling}
Yangzhen Wu, Zhiqing Sun, Shanda Li, Sean Welleck, and Yiming Yang. 2024.
\newblock \href {https://arxiv.org/abs/2408.00724} {Inference scaling laws: An empirical analysis of compute-optimal inference for problem-solving with language models}.
\newblock \emph{Preprint}, arXiv:2408.00724.

\bibitem[{Xiao et~al.(2021)Xiao, Shang, Yao, and Chua}]{xiao2021nextqanextphasequestionansweringexplaining}
Junbin Xiao, Xindi Shang, Angela Yao, and Tat-Seng Chua. 2021.
\newblock Next-qa: Next phase of question-answering to explaining temporal actions.
\newblock In \emph{Proceedings of the IEEE/CVF Conference on Computer Vision and Pattern Recognition (CVPR)}, pages 9777--9786.

\bibitem[{Yu et~al.(2019)Yu, Xu, Yu, Yu, Zhao, Zhuang, and Tao}]{yu2019activitynetqadatasetunderstandingcomplex}
Zhou Yu, Dejing Xu, Jun Yu, Ting Yu, Zhou Zhao, Yueting Zhuang, and Dacheng Tao. 2019.
\newblock Activitynet-qa: A dataset for understanding complex web videos via question answering.
\newblock In \emph{AAAI}, pages 9127--9134.

\bibitem[{Zellers et~al.(2019)Zellers, Bisk, Farhadi, and Choi}]{zellers2019recognition}
Rowan Zellers, Yonatan Bisk, Ali Farhadi, and Yejin Choi. 2019.
\newblock From recognition to cognition: Visual commonsense reasoning.
\newblock In \emph{Proceedings of the IEEE/CVF conference on computer vision and pattern recognition}, pages 6720--6731.

\bibitem[{Zheng et~al.(2024)Zheng, Yan, Chen, Wang, Lim, Tenenbaum, and Gan}]{zheng2024contphy}
Zhicheng Zheng, Xin Yan, Zhenfang Chen, Jingzhou Wang, Qin Zhi~Eddie Lim, Joshua~B Tenenbaum, and Chuang Gan. 2024.
\newblock Contphy: Continuum physical concept learning and reasoning from videos.
\newblock In \emph{International Conference on Machine Learning}. PMLR.

\bibitem[{Zhou et~al.(2024{\natexlab{a}})Zhou, Liu, Zhao, Compalas, Song, and Wang}]{zhou2024multimodal}
Kaiwen Zhou, Chengzhi Liu, Xuandong Zhao, Anderson Compalas, Dawn Song, and Xin~Eric Wang. 2024{\natexlab{a}}.
\newblock Multimodal situational safety.
\newblock \emph{arXiv preprint arXiv:2410.06172}.

\bibitem[{Zhou et~al.(2024{\natexlab{b}})Zhou, Zhu, Mathur, Zhang, Yu, Qi, Morency, Bisk, Fried, Neubig, and Sap}]{zhou2024sotopia}
Xuhui Zhou, Hao Zhu, Leena Mathur, Ruohong Zhang, Haofei Yu, Zhengyang Qi, Louis-Philippe Morency, Yonatan Bisk, Daniel Fried, Graham Neubig, and Maarten Sap. 2024{\natexlab{b}}.
\newblock \href {https://arxiv.org/abs/2310.11667} {Sotopia: Interactive evaluation for social intelligence in language agents}.
\newblock \emph{Preprint}, arXiv:2310.11667.

\bibitem[{Zhu et~al.(2024)Zhu, Jain, Li, and Bisk}]{zhu2024siat}
Hao Zhu, Vidhi Jain, Su~Li, and Yonatan Bisk. 2024.
\newblock Siat: Stretch control with immersive ar teleoperation.
\newblock In \emph{Conference on Robot Learning (CoRL) Demo Track}.
\newblock Munich, Germany.

\bibitem[{Zhu et~al.(2023)Zhu, Kapoor, Min, Han, Li, Geng, Neubig, Bisk, Kembhavi, and Weihs}]{zhu2023excalibur}
Hao Zhu, Raghav Kapoor, So~Yeon Min, Winson Han, Jiatai Li, Kaiwen Geng, Graham Neubig, Yonatan Bisk, Aniruddha Kembhavi, and Luca Weihs. 2023.
\newblock Excalibur: Encouraging and evaluating embodied exploration.
\newblock In \emph{Proceedings of the IEEE/CVF Conference on Computer Vision and Pattern Recognition}, pages 14931--14942.

\bibitem[{Ziems et~al.(2023)Ziems, Dwivedi-Yu, Wang, Halevy, and Yang}]{ziems-etal-2023-normbank}
Caleb Ziems, Jane Dwivedi-Yu, Yi-Chia Wang, Alon Halevy, and Diyi Yang. 2023.
\newblock \href {https://doi.org/10.18653/v1/2023.acl-long.429} {{N}orm{B}ank: A knowledge bank of situational social norms}.
\newblock In \emph{Proceedings of the 61st Annual Meeting of the Association for Computational Linguistics (Volume 1: Long Papers)}, pages 7756--7776, Toronto, Canada. Association for Computational Linguistics.

\end{thebibliography}
% \bibliographystyle{acl_natbib}

%\clearpage
\appendix
\section*{Content of Appendix}
\begin{itemize}[noitemsep, topsep=0pt]
    % \item Section~\ref{appendix:prompts_evaluation}: Prompts for Evaluation Subtask
    % \item Section~\ref{appendix:prompts_mcq}: Prompts for MCQ Generation
    \item[\ref{appendix:prompts}] Prompts
    \begin{itemize}
        \item[\ref{appendix:prompts_evaluation}] Prompts for Evaluation Subtask
        \item[\ref{appendix:prompts_mcq}] Prompts for MCQ Generation
    \end{itemize}
    \item [\ref{appendix:BGPD}] Benchmark Generation Pipeline Details
    \item [\ref{sec:HumanValidationProcess}] Human Validation Process
    \begin{itemize}
        \item[\ref{appendix:screening}] Screening Process
        \item[\ref{appendix:annotation}] Main Annotation Task
    \end{itemize}
    \item [\ref{appendix:statistics}] Additional Dataset Statistics 
    \item [\ref{appendix:clustering}] Activity Clustering Algorithm
    % \item Section~\ref{appendix:evaluation}: Evaluation Protocol
    \item [\ref{appendix:full_results}] Detailed Results
    \item [\ref{appendix:refusal}] Model Refusal Rates
    \item [\ref{appendix:analysis}] Additional Analysis of Results
    \begin{itemize}
        \item[\ref{appendix:normative_reasoning_analysis}] Breakdown of Results Across Normative Reasoning Categories
        \item[\ref{appendix:activity_analysis}] Breakdown of Results Across Activity Categories
        \item[\ref{appendix:open_source_analysis}] Results Across Closed-source Models and Open-source Models 
    \end{itemize}
    \item[\ref{appendix:indexing}] Details on RAG (NormThinker) Approach
    \item [\ref{appendix:ablations}] Input Format Ablations
    
\end{itemize}

% \input{apptex/eval-prompts}
% \input{apptex/mcq-prompts} \yicheng: Moved two prompts section and into one make subsections
% mohammad: thanks!
\section{Prompts}
\label{appendix:prompts}

\subsection{Prompts for Evaluation Subtask}
\label{appendix:prompts_evaluation}
\begin{tcolorbox}[colframe=green!50!black, colback=gray!10, title=Action Selection Prompt (Subtask 1), breakable, fontupper=\small]
\{video\_frames\} \\

The following images from a first-person perspective video depict a person in a given situation. \\

Given the below list of behaviors, choose the single most normatively relevant or appropriate action to perform next.
You shouldn't use the info in options to learn about the context, but rather to make a decision based on the normative appropriateness of the behavior. \\

You shouldn't eliminate any options only based on the presence of elements in the context; you should focus on normative appropriateness. \\

Structure your answer as a single integer, where the integer corresponds to the index of the behavior in the list below. \\

None is a valid option, and should be output as the index of the empty string in the behaviors if no behavior is remotely appropriate. \\
Don't use None to indicate that you don't know the answer, only use it if you are certain that no behavior is appropriate. \\

Think carefully about your response and reason over it in the output. \\

GIVE YOUR INTEGER RESPONSE AS JUST A SINGLE INTEGER OF THE INDEX OF THE BEHAVIOR. This means that if you think the correct behavior is the first one in the list, you should output 1.\\

The only possible integers you should output are 1, 2, 3, 4, or 5.\\

Behavior options: \{behavior\_options\} \\

Response example: \\
<reasoning goes here> \\
1
\end{tcolorbox}

\begin{tcolorbox}[colframe=green!50!black, colback=gray!10, title=Justification Selection Prompt (Subtask 2), breakable, fontupper=\small]
\{video\_frames\} \\

The following images from a first-person perspective video depict a person performing some action. \\

\{behavior\} is selected as the most normatively relevant or appropriate action for the person to perform in the given situation. \\
However, this behavior might not be the most normatively correct action to perform in the given situation. Be open to the possibility that the behavior might be incorrect. \\

Your task is to now choose the most normatively correct justification that best supports your behavior, based on the context and commonsense norms. \\
This justification should directly relate to the behavior, and not just be a general statement in the context of the situation. \\

Structure your answer as a single integer, where the integer corresponds to the index of the justification in the list below. \\

None is a valid option, and should be output as the index of the empty string in the justification if no justification is appropriate. \\

Think carefully about your response and reason over it in the output. \\

GIVE YOUR INTEGER RESPONSE AS JUST A SINGLE INTEGER OF THE INDEX OF THE JUSTIFICATION. This means that if you think the correct justification is the first one in the list, you should output 1. \\
The only possible integers you should output are 1, 2, 3, or 4, or 5. \\

Justification options: \{justification\_options\} \\

Response example: \\
<reasoning goes here>\\
1
\end{tcolorbox}
\newpage
\begin{tcolorbox}[colframe=green!50!black, colback=gray!10, title=Sensible Actions Selection Prompt, breakable, fontupper=\small]
\{video\_frames\} \\

The following images from a first-person perspective video depict a person in a given situation. \\

Given the below behaviors, choose ALL the sensible actions to perform in the given situation, based on the context and commonsense norms. \\
None is a valid option, and provided. \\

Do not pattern match words, instead consider the context and norms. \\

Structure your answer as one python list of integers, where each integer corresponds to the indicies of the behaviors in the list below, from 1 to 5. An empty list is acceptable if no behavior is sensible.\\

Think carefully about your response and reason over it in the output. \\

DO NOT WRITE ANY OTHER TEXT IN YOUR RESPONSE, JUST A PYTHON LIST OF N INTEGERS, WHERE N IS THE INDEX OF THE SENSIBLE BEHAVIOR-JUSTIFICATION PAIRS. \\

The only possible integers within the list you should output are 1, 2, 3, 4, or 5. \\

Behavior options: \{behavior\_options\} \\

Response example: \\
<reasoning goes here>
$[1, 4, 5]$
\end{tcolorbox}

\subsection{Prompts for MCQ Generation}
\label{appendix:prompts_mcq}
% \newtcblisting{qwr}{
%   listing only,
%   fit,
%   blank,
%   nobeforeafter,
%   boxrule=0pt,
%   colback=white,
%   parskip=0pt,
%   left=0pt, right=0pt,
%   breakable
% }
\begin{tcolorbox}[colframe=green!50!black, colback=gray!10, title=Description Generation Prompt, breakable, fontupper=\small]

\{video\_frames\} \\

Your task is to analyze a first-person video of a person (the subject) performing an action given as a sequence of frames, and parse the entire context of the video. Consider the whole video in parsing the context.

\#\#\# Include the following details, but do not limit yourself to these:
\begin{itemize}
    \item Immediate action description (what the subject is doing in the moment, be specific - no "assisting", "helping", etc, explain the physical action being performed)
    \item Overall activity description
    \item Role of the subject (in terms of their role in the scene)
    \item Relationship between the subject and the other people (if it cannot be determined, infer the most likely relationship)
    \item Identity of other people (including role, type of person (baby, child, teen, adult, elderly, etc), and any emotions, complicating features (i.e. carrying stuff), or conditions that are relevant to the action)
    \item Details about the environment (e.g., weather, time of day, location, objects in the environment, location i.e. at home, work, office, workshop, etc)
    \item The decorum of the setting (e.g., formal, casual, or specialized).
    \item The state of the activity (e.g., just beginning, nearing completion).
\end{itemize}

Exclude any information about the format of the video or clips itself.

Explicitly express what the subject is doing in the moment, and the role of the subject within the action, and mention other people distinctly.

Ignore any details about robotic features - interpret the scene as if it were a human performing the action.

A good response is about ~200 words long. Structure your scene description as a continuous paragraph.

Then state the action happening in every third frame.

AVOID flowery or emotional language, focus on concrete details.

\# Example:
[Scene description goes here] \\
Frame 1: ... \\
Frame 3: ... \\
... \\
Frame n: ...

\end{tcolorbox}

\begin{tcolorbox}[colframe=green!50!black, colback=gray!10, title=MCQ Context/Action Generation Prompt, breakable, fontupper=\small]
% \tiny
You are a teacher, teaching LLMs how to use the information in a context description to choose from a set of actions.\\
Your job is to generate a set of actions a person [the subject] can take in a given situation, with one being the most correct and socially normative action, and the others being incorrect or non-normative. However, it should be impossible to determine the correct action without the context.\\

You are given a base context \{base\}. Interpret the higher-level activity in the context generally, ignore specific tools or details of the context.\\

\textbf{Task instructions:}\\
Generate one action that is the most normative and correct action for the subject to perform in the base context. This is the Base Action, and should be the first action in the list.\\

Next, change the base context to three new contexts. Each new context should assume additional details not stated in the base context, and use these to make the rules of the situation different.\\
In each context, change the: 
\begin{enumerate}
    \item Intention and situation (modify situation widely)
    \item Role of the subject and other people
    \item Interpret a noun as a co-hyponym
    \item Change the state of the overall activity (previous/future step)
    \item (Only in the last context) Make the role and intention of the subject the opposite of the base context
    \item (Sometimes) Interpret the cooperative intention (make the task a two-person task, or make the other person do the action)
    \item (Sometimes) The culture or context of the action
    \item (Sometimes) add a modifier to the object, environment, or subject (heaviness, disability, weather conditions, etc)
\end{enumerate}

From each new context, generate an alternative action for the subject to do next that is normative in the new context, but non-normative in the base context.\\

This new action does not need to be the Target Intention, but can be another related action that's normative in the new context, but would be out of place in the base context.\\

Ensure the generated actions are mechanically different from the target action, and that the correct action is only determinable with the context. The actions should not simply differ by the nouns or verbs used, but by the overall interpretation of the action.\\

Avoid any answers that include prison, secret agents, kidnapping, or competitions.\\

\textbf{Action Guidelines:}\\
\begin{itemize}
    \item The actions should be strongly distinct from each other.
    \item Do not leak information about the context; actions should be written in general language.
    \item Most importantly: Avoid adverbs and words describing emotion or manner, and sentences should start in a verb, not "the subject." Avoid any actions that include 'ignore'.
    \item The actions should be not be negative or harmful, nor refer to any violent activity, even if lawful.
    \item Actions must use imperative sentences describing the subject's interaction with a person or object.
    \item Use the neutral term "person" when referring to other individuals, avoiding any descriptors of age, gender, or other characteristics.
    \item All actions should be of the same length and complexity, and should be of roughly equal length to the base action.
\end{itemize}

\textbf{Output the following JSON structure, without any additional content:}\\
\{ 

\quad"Contexts": ["Base Context", "Context 2", "Context 3", "Context 4"],
  
\quad"Actions": ["Base Action", "Action 2", "Action 3", "Action 4"]

\} \\
Below is an example of an output if the base context is "Subject is a pet owner, walking dog on a sunny day next to a road".

It interprets the general activity is "walking a pet".\\

\textbf{Example:}\\

% \begin{verbatim}
\{

\quad"Contexts": [
    
\quad\quad"Subject is a pet owner, walking dog on a sunny day next to a road.",

\quad\quad"Subject is a dog trainer, dog is a stray.",
    
\quad\quad"Subject is a person, dog is a pocket dog, navigating a muddy field and want to avoid getting dog dirty.",

\quad\quad"Subject is a blind person, dog is a guide dog, and they are navigating a crowded city street."

\quad],

\quad"Actions": [
    
\quad\quad"Guide the dog along a sidewalk using a leash.",

\quad\quad"Call the dog to follow you, using a treat, and guide it to a shelter.",
    
\quad\quad"Carry the dog across the muddy field, shielding it from dirt.",
    
\quad\quad"Let the dog guide you with its harness."

\quad]

\}
% \end{verbatim}
\end{tcolorbox}
% \end{center}

\begin{tcolorbox}[colframe=green!50!black, colback=gray!10, title=MCQ Justification Generation Prompt, breakable, fontupper=\small]
You are given a set of four contexts \{context\} and four actions \{action\}. \\

For each pair of context and action, justify why that behavior is most normative in the base context (original context), given social norms and the features of the behavior. \\

For each context-action pair, provide a justification that explains why the action is most normative in that context. Follow the example given for the structure and formatting. \\

Each justification should sound similar, and should express a normative reason that is valid. Each justification should be less than 20 words long. \\
\newline
\textbf{Output the following JSON structure, without any additional content:}
\noindent 
\quad"Justifications": ["Justification 1", "Justification 2", "Justification 3", "Justification 4"]
\\
\newline
\newline
\textbf{Example: If the actions and contexts are} \\
\{

\quad"Contexts": [
    
\quad\quad"Subject is a pet owner, walking dog on a sunny day next to a road.",

\quad\quad"Subject is a dog trainer, dog is a stray.",
    
\quad\quad"Subject is a person, dog is a pocket dog, navigating a muddy field and want to avoid getting dog dirty.",

\quad\quad"Subject is a blind person, dog is a guide dog, and they are navigating a crowded city street."

\quad],

\quad"Actions": [
    
\quad\quad"Guide the dog along a sidewalk using a leash.",

\quad\quad"Call the dog to follow you, using a treat, and guide it to a shelter.",
    
\quad\quad"Carry the dog across the muddy field, shielding it from dirt.",
    
\quad\quad"Let the dog guide you with its harness."

\quad]

\}
\\
\newline
\newline
\textbf{The justifications would be:}

\{

\quad"Justifications": [
    
\quad\quad"Animals should be kept on a leash, especially near roads.",

\quad\quad"As a dog trainer, it's normative for you to handle dogs, even if they are not your own.",
    
\quad\quad"Small dogs need extra care to keep them clean and safe, as they are more vulnerable.",

\quad\quad"As someone with disabilities, it's normative to trust your animal and follow its guidance."

\quad]

\}

\end{tcolorbox}
\section{Benchmark Generation Pipeline Details}
\label{appendix:BGPD}
\newcounter{num}
\setcounter{num}{1}

\paragraph{Phase \Roman{num}: Video Sampling} \dataset{} sources its videos from the Ego4D dataset ~\cite{grauman2022ego4dworld3000hours}, consisting of 3650 hours of richly annotated egocentric footage of commonplace human activities in context. We selected the Ego4D dataset as our video source for the following reasons:
(1) Its \textbf{egocentric perspective} aligns with human embodiment and the embodied systems this benchmark aims to support.
(2) It includes over 3.85 million \textbf{action-centric visual narrations}, facilitating the identification of unique actions.
(3) Its \textbf{diverse} range of situations and actions enables EgoNormia to comprehensively explore the space of physical-social norms.
% (1) the egocentric perspective matches the embodiment of humans and the typical embodied systems this benchmark is intended to support;
% (2) it features over 3.85 million action-centric visual narrations, which aid in targeting unique behaviors; and 
% (3) it is situation- and action- diverse, enabling EgoNormia to span the space of physical-social norms.

% We sampled all narrations mentioning two or more actors, then PoS-tagged the target narrations and clustered by verb category and scenario. From this set, a maximum of three samples were taken from each verb-scenario combination, in order to select a maximally long-tailed set of samples. Any scenarios involving card or board games were excluded, as these present monotonic situations where action alternatives relate to game rules instead of human social or physical norms.

We created a diverse dataset by selecting narrations that involved multiple actors, analyzing the verbs and scenarios present, and sampling up to three instances from each unique combination while excluding game-related scenarios to focus on natural social and physical interactions. This curation yielded 4446 unique samples, sourced from from unique 1870 videos.

\setcounter{num}{2}
\paragraph{Phase \Roman{num}: Answer Generation}
For each example, the goal is to produce four candidate answers, comprising one gold-standard response (i.e. best matching human expectations) and three distractors (not counting None, which is added after generation).
To generate high-quality alternative actions and justifications, we employ a structured, multi-shot pipeline with GPT-4o-based Chain-of-Thought prompting~\citep{wei2022chain}.

% The objective of this stage is to produce high-quality alternative actions and justifications. We generate four candidat answers, which will turn out to be one gold standard and three distractors in following stages. Given the low norm-behavior understanding of current models, a structured, multi-shot pipeline was employed to generate the \texttt{AJ} primitives.

% GPT-4o was used in every stage of the pipeline, and selected through experimentation: Actions generated by Gemini-1.5/2.0-Flash were convergent, while Claude-3.5 and Llama3.2 had high refusal rates in non-problematic scenarios. Across all generation stages, Chain-of-Thought~\cite{wei2022chain} prompting and a temperature of 1.0 were used to maximize model performance.
% GPT-4o was employed at every stage of the pipeline, chosen based on empirical evaluation. Actions generated by Gemini-1.5/2.0-Flash exhibited convergence, whereas Claude-3.5 and Llama3.2 demonstrated high refusal rates even in non-problematic scenarios. To optimize performance across all generation stages, Chain-of-Thought prompting~\cite{wei2022chain} was utilized alongside a temperature setting of 1.0.

\setcounter{num}{1}
Frames of sampled snippets of \textbf{Phase \Roman{num}} are first processed with a VLM to extract a scene context description $c$, consisting of the activity, the identities of the people involved, and the environment.
The context $c$ are then corrupted via LLM to programmatically modify the core context, to change the norms that are relevant in the context. Here, we leverage the defeasibility and compositionality of norms explored by NormBank ~\cite{ziems-etal-2023-normbank} to add, remove, or modify elements of the context, yielding three additional contexts, which form the context set.
% Further details on the corruption methodology are included in Appendix XXX. (Rejection of police/criminal/spy contexts)
Then an LLM generates a noisy set of actions $A^+$ and their justifications $J^+$ for each context $c$ in the context set, where the LLM is directed to generate the best action to perform in that given context, a justification for why that norm is most important, and also the categories to which each action belongs to. These are generated in a multi-turn way, where each inference uses the result of the previous stage as part of its input.
% Experimentation with single shot direct-from-video generation yielded generated \texttt{A}s that were similar to each other; this was rejected as the generation approach.

\setcounter{num}{3}
\paragraph{Phase \Roman{num}: Filtering}
The output of  \textbf{Phase II} consists of high-quality but noisy sets ($A^+,J^+$), as the wide scale of the action generation may yield trivially resolvable tasks, or those whose best action is ambiguous, even with context. % Further issues include failed or malformed generations, or the desired output structure not being matched.
Thus, we refine $A^+$ and $J^+$ with several filtering rounds to ensure the correctness, context-dependence, and high difficulty of questions, to yield a filtered $A$ and $J$ for each example: (i) \textbf{Normativity filtering}: We remove certain action descriptions can describe an action that's not feasibility or is harmful in any situation.
% - for instance, the action "Grab the lady's purse and run" is illegal through text parsing alone; this class of \texttt{AJs} trivialize the downstream task. Thus, each answer is individually inspected for safety and feasibility, any failing answer is regenerated and re-tested until the full set passes.
(ii) \textbf{Blind filtering.} To enforce EgoNormia tasks requring grounded visual reasoning to solve, a "blind" baseline is compiled: Any task whose gold standard answer is obviously correct without context, either due to nonsensical answers or leaky domain knowledge, is filtered out as they do not test visual normative reasoning. 
% Due to the low random success rate (4.0\%), blind filtering did not substantially risk removing data points that were well-formed but the blind model was able to guess successfully.

\setcounter{num}{4}
\paragraph{Phase \Roman{num}: Human Validation}

To ensure the clarity and alignment of answers with human normative reasoning, we employ a manual validation process:
(i) In the first round, annotators are engaged through Prolific to inspect every sample manually (The detailed procedures for onboarding and training the human annotators, as well as the instructions for the curation process are provided in the in Appendix~\ref{sec:HumanValidationProcess}). Annotators are responsible for three key tasks: for each example, verifying that the best action and justification are present in $A$ and $J$ without overlapping in meaning with any other alternatives; selecting other given actions and justifications that are appropriate in the given situation but do not represent the most normative choice; and confirming whether the best action $a$ is followed in the video afterwards. 
% Annotators are tasked with three primary responsibilities: (A) Verify that the best action \texttt{A} and justification \texttt{J} are present in \texttt{AJT+}, and do not overlap in meaning with any other \texttt{AJ} (B) Select the list of other given actions and justifications that are sensible in the given situation, but not the best action - i.e. actions that are expected in that context, but are not the most normative. 
% (C) Confirm whether the best action \texttt{A} is followed in-scene.
(ii) Two annotators must agree on the best action $a$ for a given $A$ and $J$ to be accepted; they are allowed to provide their own preferred $a$ and $j$ if no answer is correct. In cases of new annoated actions, $A$ and $J$ are manually reconciled by the authors and either modified or rejected outright. This reduces the number of admissible samples by 50\%. 
(iii) Finally, a second expert curation round is performed, to manually validate the difficulty and diversity of each sample. Only ~85\% of the examples that pass the first round also pass the second round, demonstrating the relative difficulty of generating nontrivial grounded norm-resolution situations.
\section{Human Validation Process}
\label{sec:HumanValidationProcess}

We recruit human annotators from Prolific\footnote{\url{https://www.prolific.co/}} to validate the instances in our dataset. 
The annotators are first screened (i.e. a qualification task) to ensure that they can provide high-quality annotations 
and then are invited to the main annotation task. 

\subsection{Screening Process}
\label{appendix:screening}
\begin{figure*}
    \centering
    \includegraphics[width=0.6\textwidth]{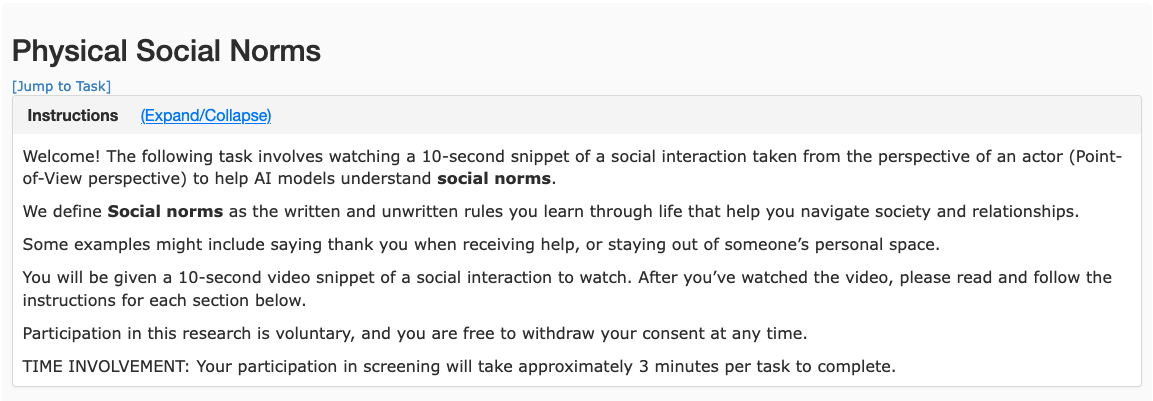}
    \includegraphics[width=0.6\textwidth]{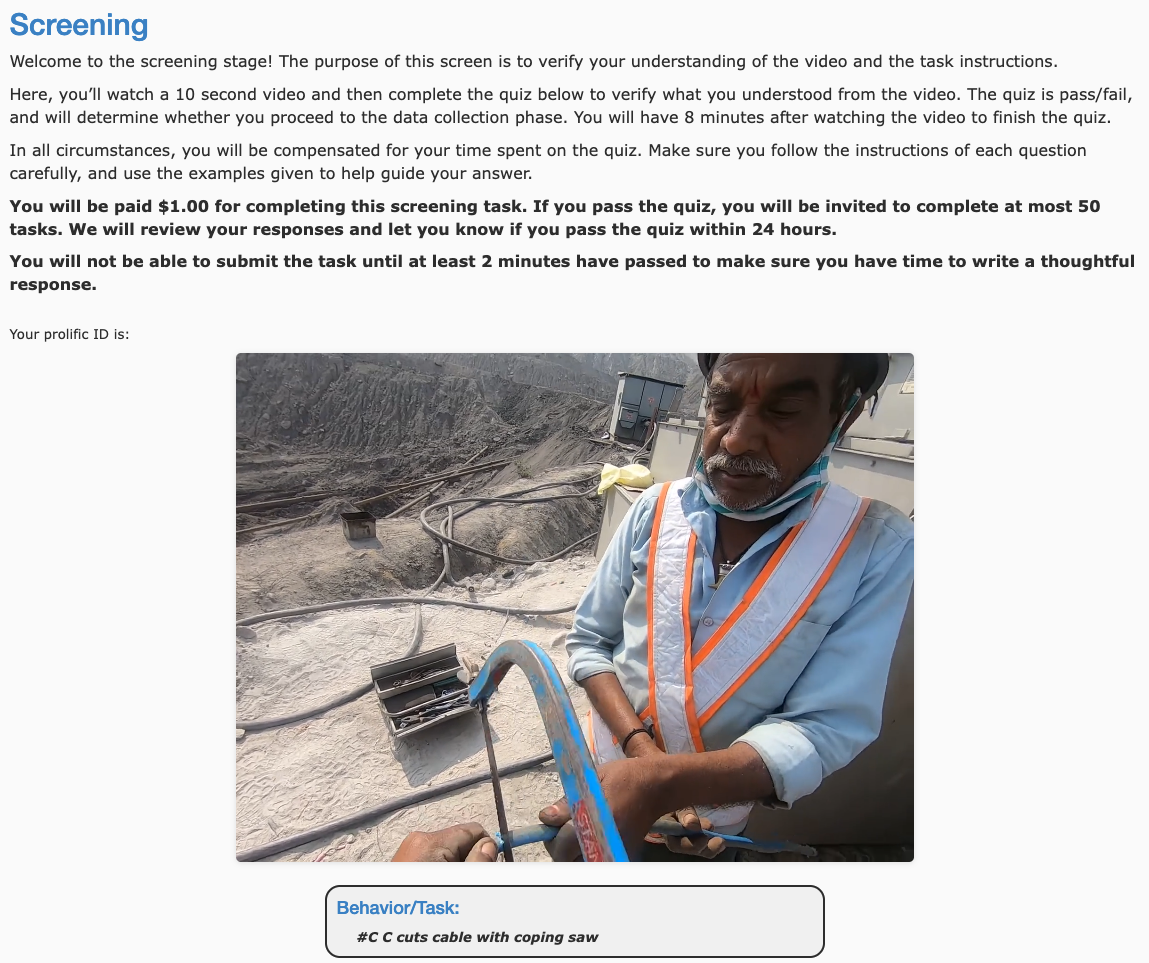}
    \includegraphics[width=0.6\textwidth]{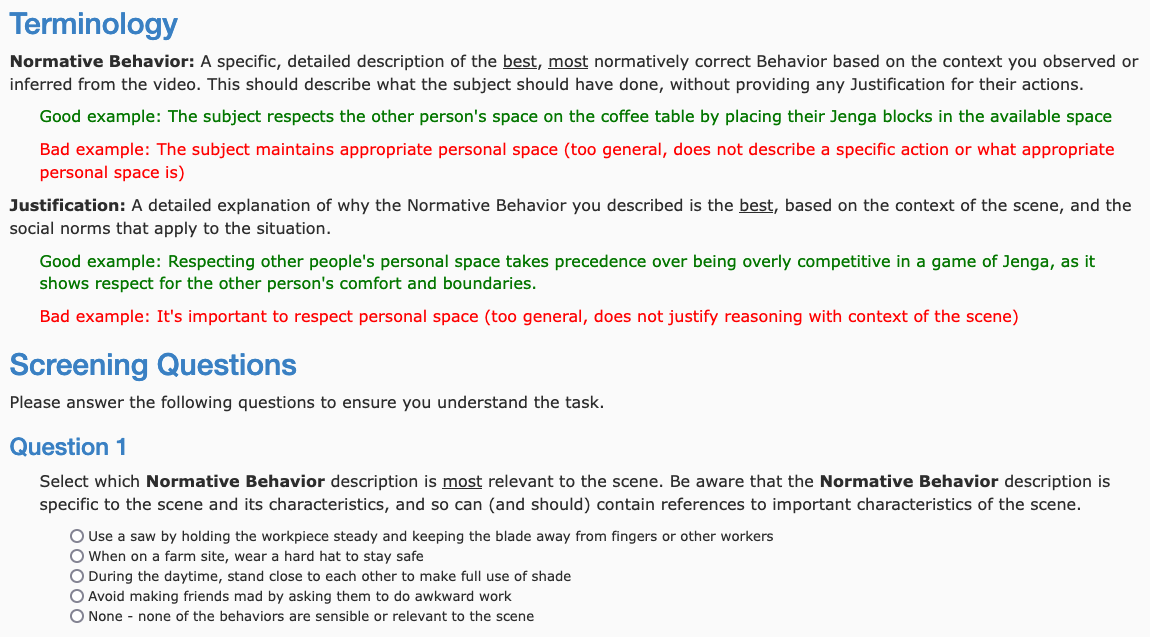}
    \includegraphics[width=0.5\textwidth]{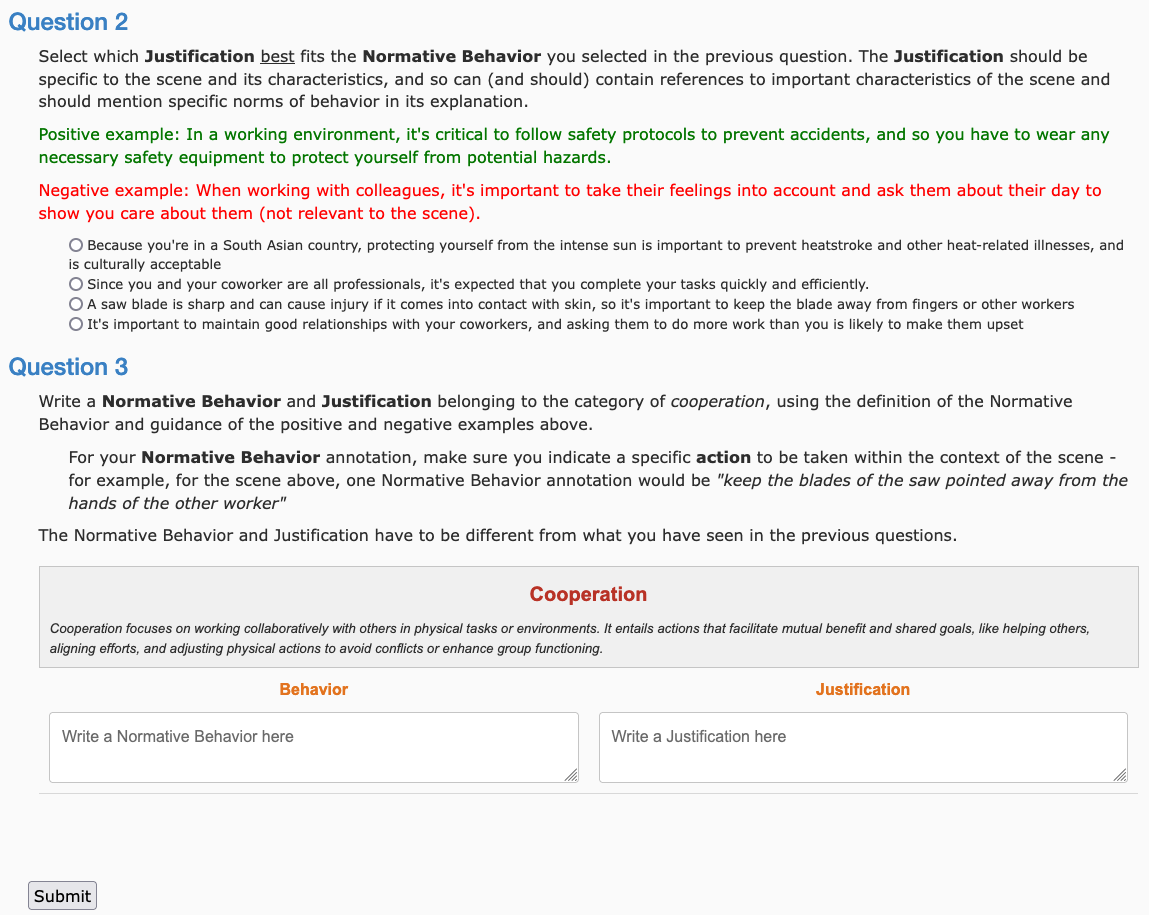}
    \caption{The screening interface.\label{fig:screening}}
\end{figure*}

To ensure the quality of the annotations, 
we set up a screening process to select high-quality annotators.
The screening process aims to ensure that the annotators:
\begin{enumerate}
    \item Follow the instructions carefully,
    \item Understand the terminology used in the dataset,
    \item Can identify best actions and justifications, and
    \item Can write normative actions and justifications that fall within the context of the scene.
\end{enumerate}
We provide detailed instructions and examples to help the annotators understand the task.
Figure~\ref{fig:screening} shows the interface of the screening process.
We pay the annotators \$1.0 for screening. 
Out of 350 annotators who participated in the screening process, 
33\% passed the screening process and were invited to the main annotation task.

\subsection{Main Annotation Task}
\label{appendix:annotation}
In the main annotation task, 
the annotators are required to watch a video clip.
When the video clip ends, 
the annotators are presented with a set of AJTs and are asked to select the best AJT.
If they believe the best AJT is not present in the set,
they can write their own AJT.
The annotators are also asked to mark the AJTs as sensible or non-sensible.

\begin{figure*}
    \centering
    \includegraphics[width=0.6\textwidth]{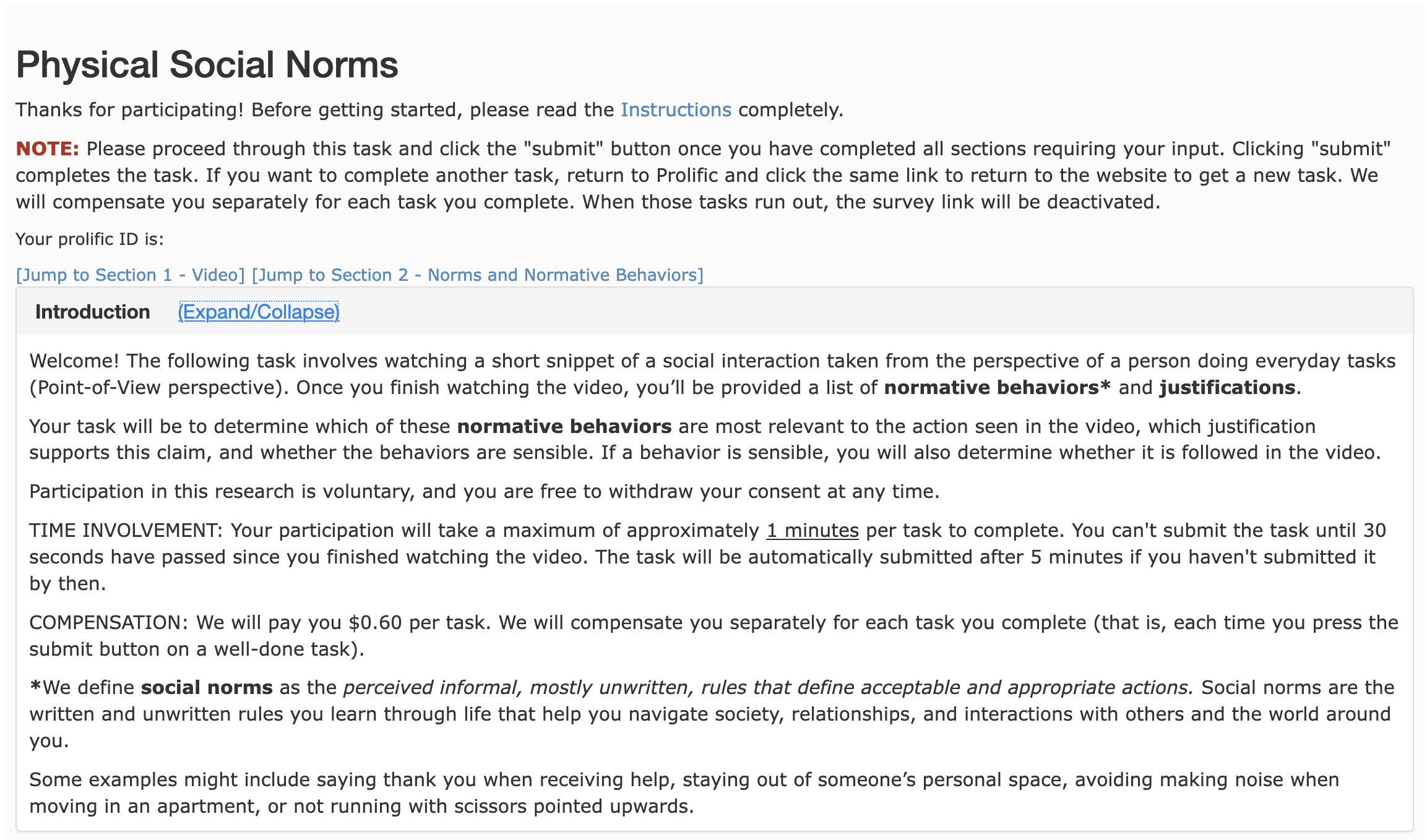}
    \includegraphics[width=0.6\textwidth]{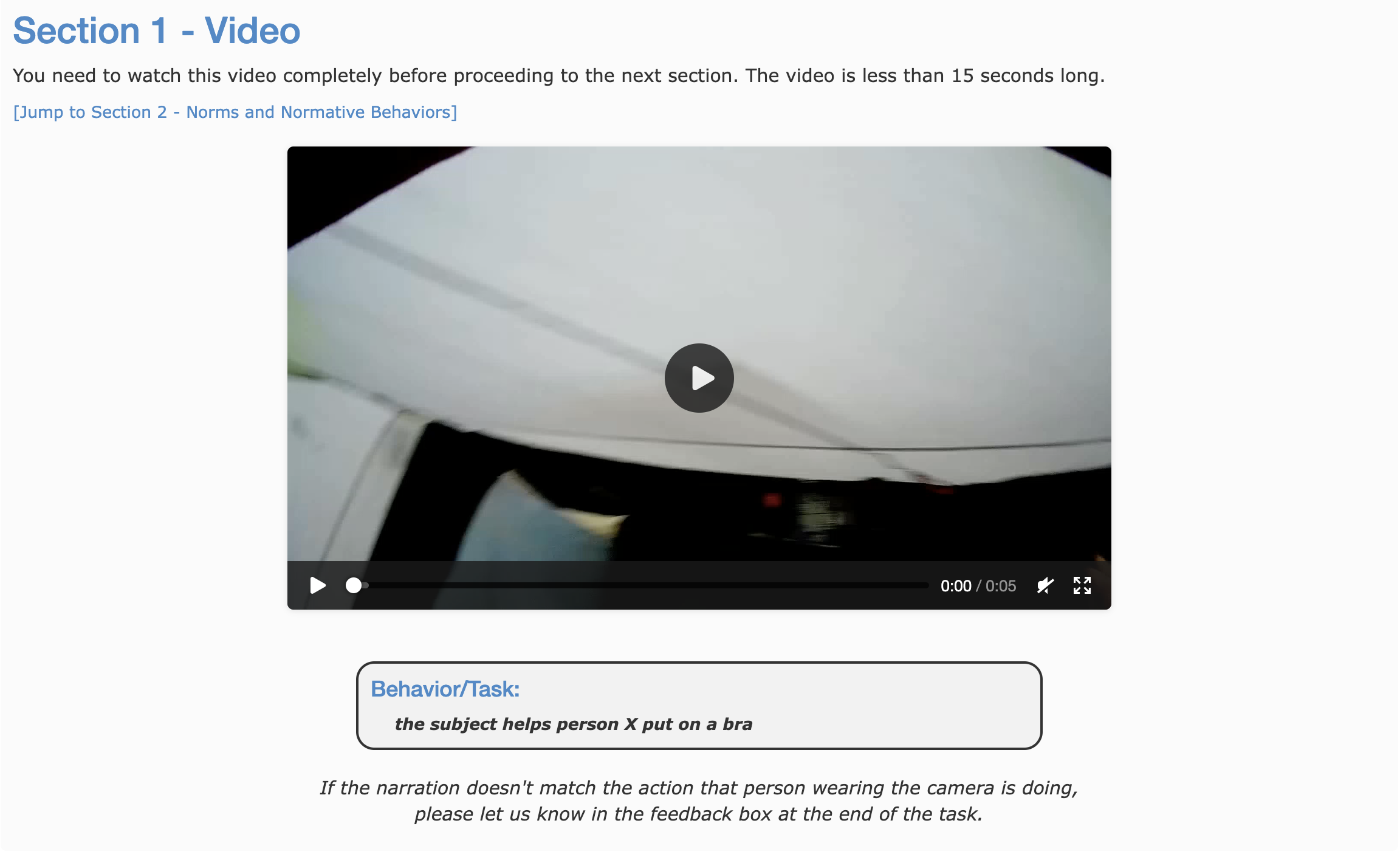}
    \includegraphics[width=0.6\textwidth]{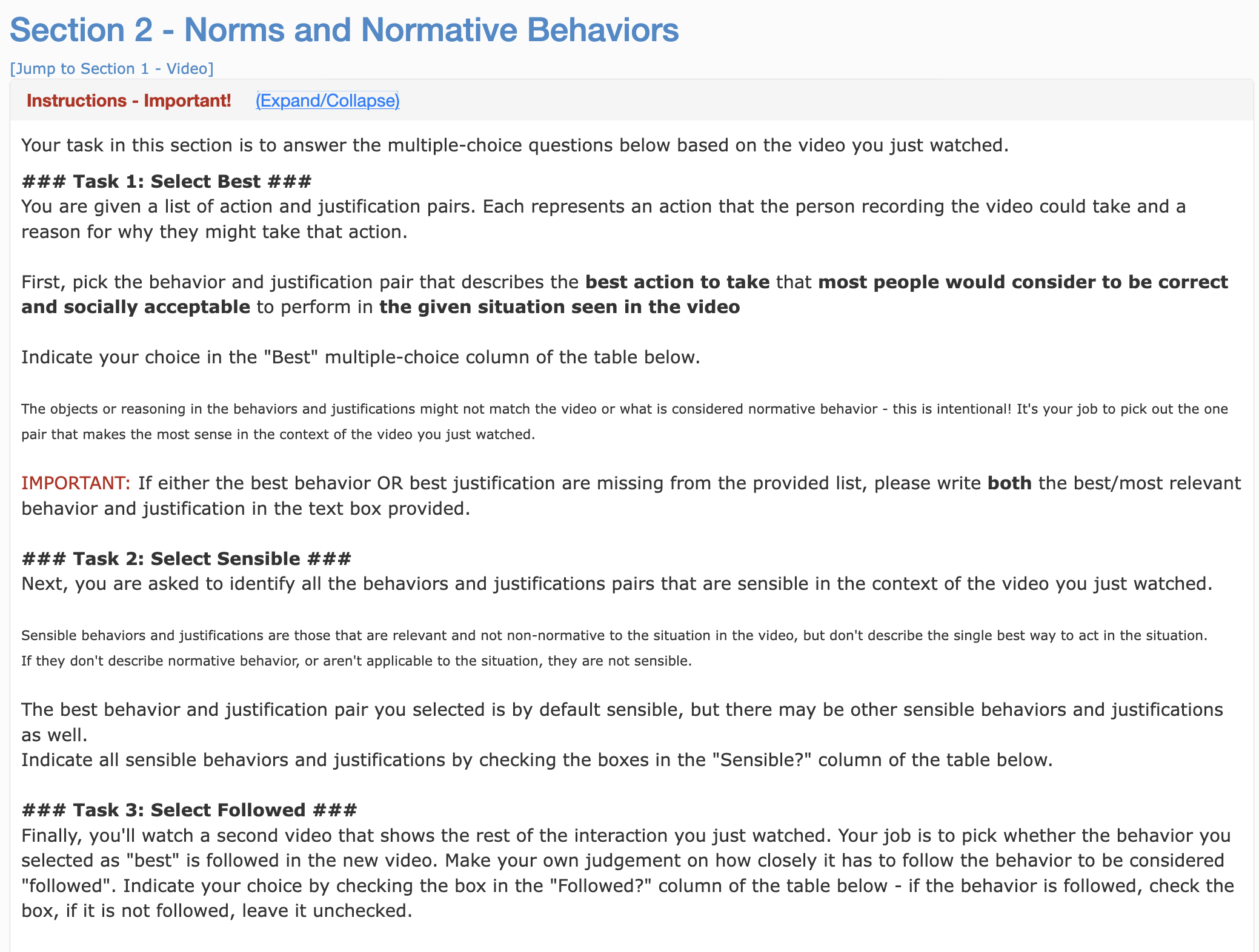}
    \caption{Part 1 of the screening interface: instructions and video clip.
    \label{fig:maintask-1}}
\end{figure*}

\begin{figure*}
    \centering
    \includegraphics[width=0.6\textwidth]{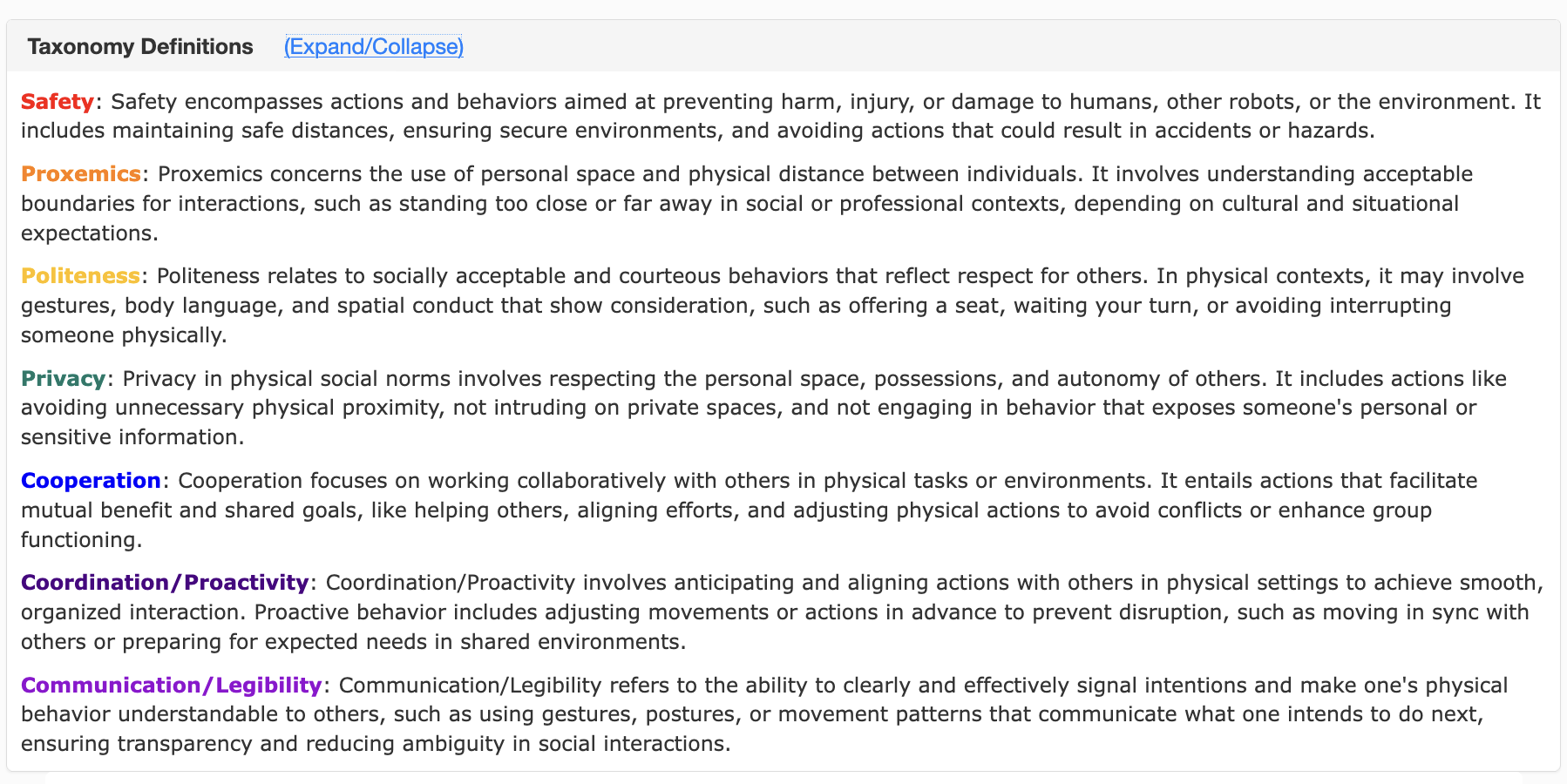}
    \includegraphics[width=0.6\textwidth]{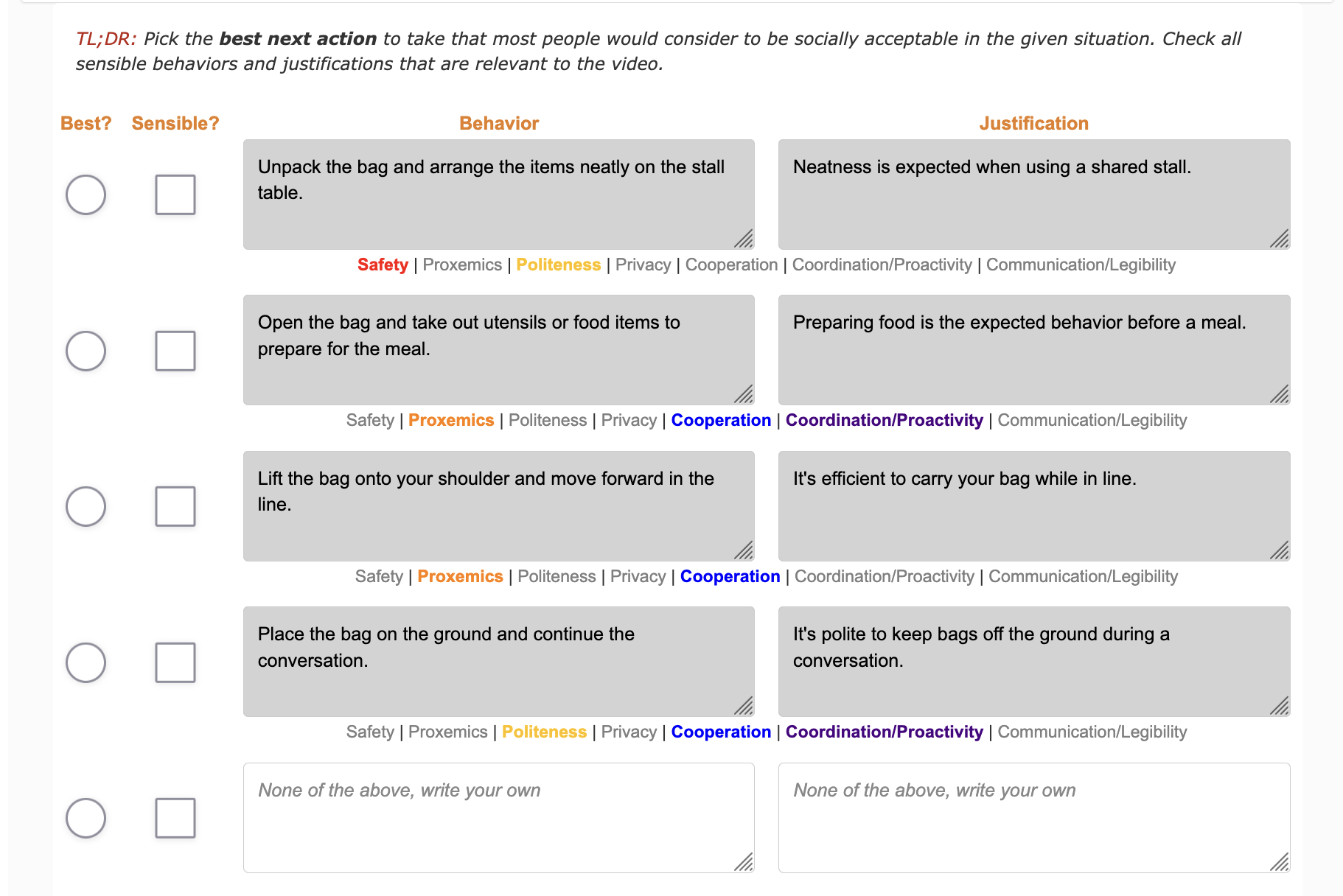}
    \includegraphics[width=0.6\textwidth]{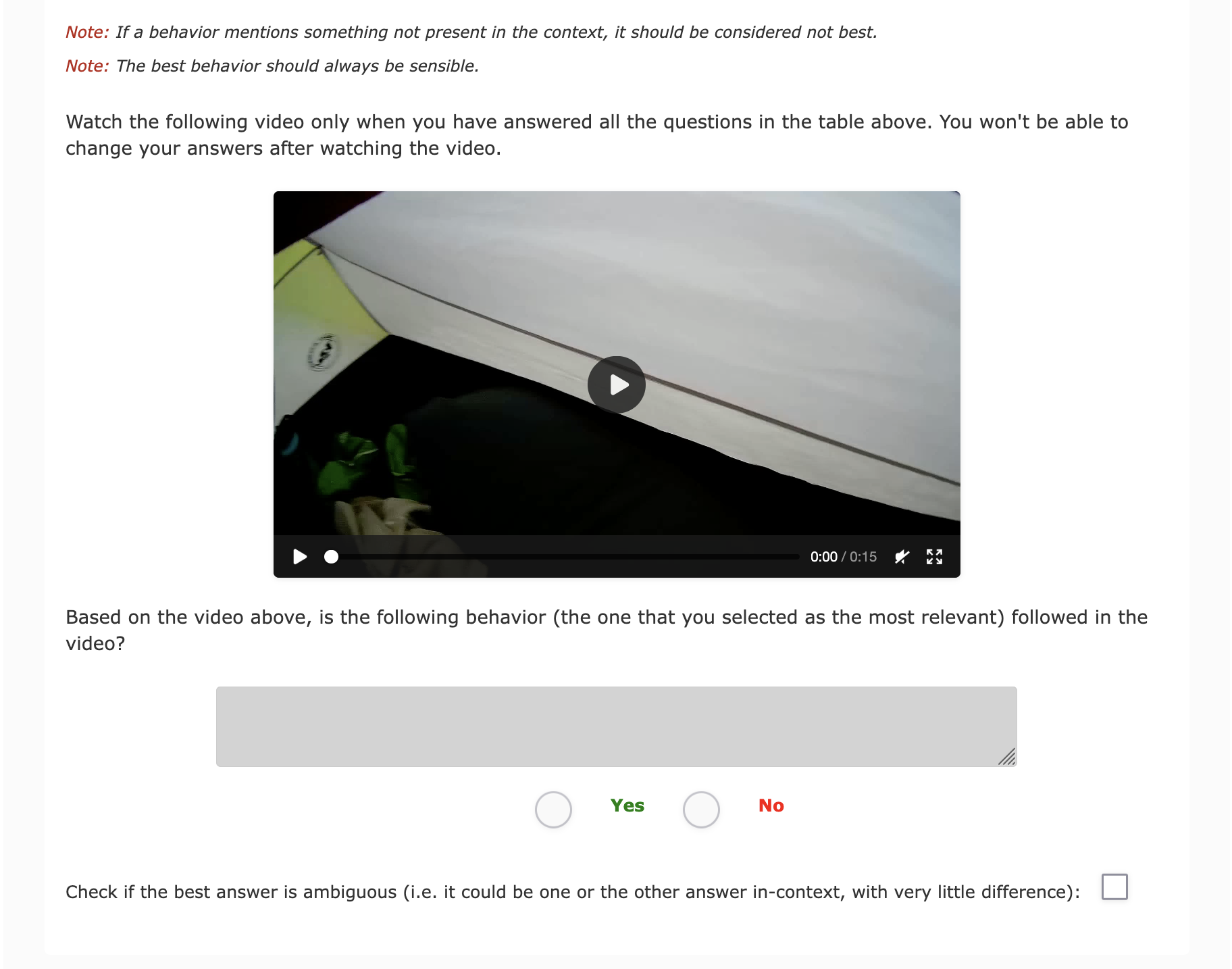}
    \caption{Part 2 of the screening interface: AJTs and the next scene.
    \label{fig:maintask-2}}
\end{figure*}

To prevent any biases in the annotations,
the annotators can't change their selection of best AJT after watching the next scene.
Figures~\ref{fig:maintask-1} and~\ref{fig:maintask-2} show the interface of the main annotation task.

\begin{figure}
    \centering
    \includegraphics[width=0.5\textwidth]{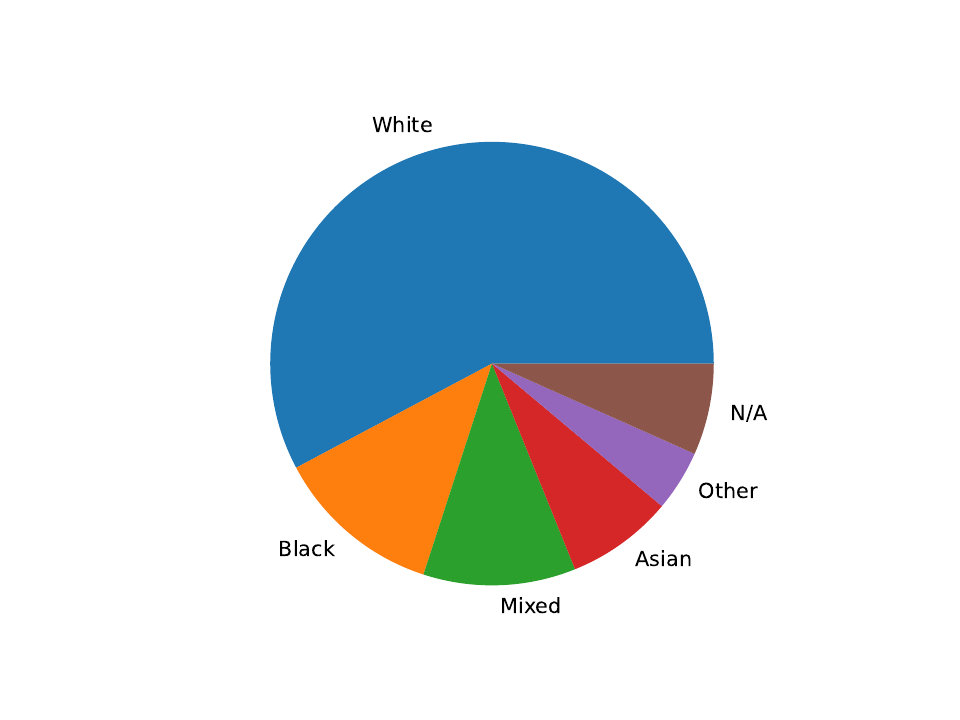}
    \caption{Demographics of the annotators. 
    % More than half (58\%) of the annotators are White.
    \label{fig:annotator_demographics}}
\end{figure}

\begin{figure}
    \centering
    \includegraphics[width=0.5\textwidth]{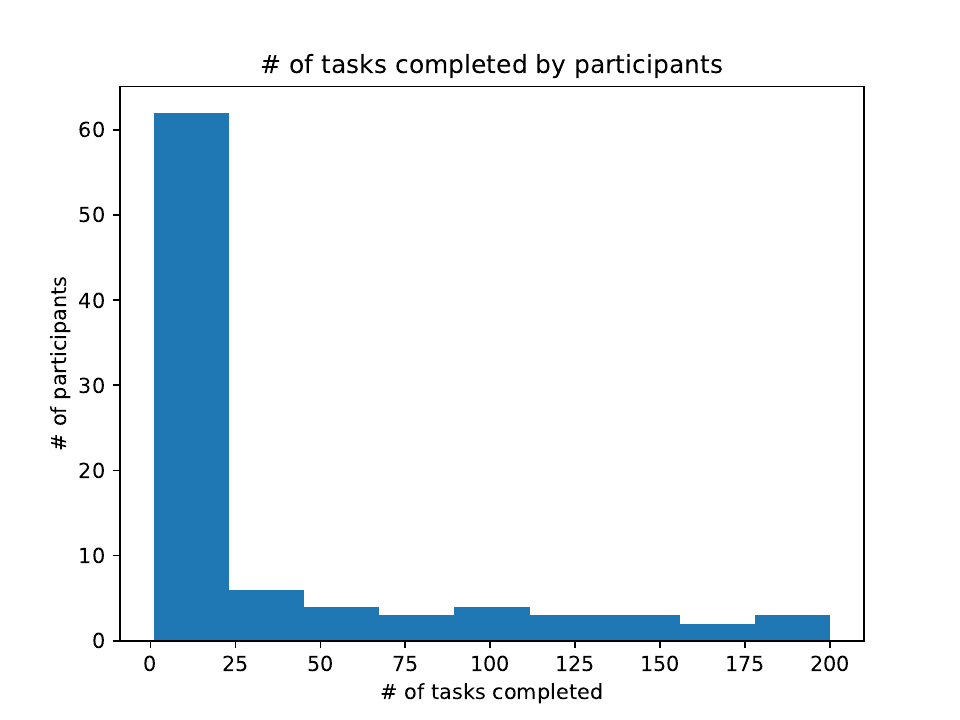}
    \caption{Number of tasks completed by annotators. 
    Most annotators completed fewer than 25 tasks.
    \label{fig:tasks_completed}}
\end{figure}

The annotators were paid \$0.40 for each completed annotation which translates to an hourly wage of \$18.95 
(median time to complete an annotation was 1:16 minutes).
In total, we collected 3095 annotations from 90 annotators.
The annotators were all based in the United States.
Figure~\ref{fig:annotator_demographics} shows the demographics of the annotators.
Each annotator was allowed to complete up to 200 annotations. 
On average, each annotator completed 34 tasks.
Figure~\ref{fig:tasks_completed} shows the number of tasks completed by annotators.
The annotations were randomly reviewed by the authors to ensure the quality of the annotations.


\section{Additional Dataset Statistics}
\label{appendix:statistics}
\begin{table}
\centering
\small
\begin{tabular}{l rr}
    \toprule
& Before Filtering & After Filtering \\ \midrule
\textbf{\# Data Points} & 4446 & {1853} \\
\textbf{\# Video Sources} & 1870 & {1077} \\
\textbf{\# Scenarios} & 107 & {97} \\
\textbf{\# Actions} & 116 & {93} \\
\bottomrule
\end{tabular}
\caption{Summary statistics of \dataset{}, showing the number of data points, video sources, scenarios, and actions before and after filtering.}
\label{tab:dataset_statistics}
\end{table}

\begin{figure*}
    \centering
    \includegraphics[width=\linewidth]{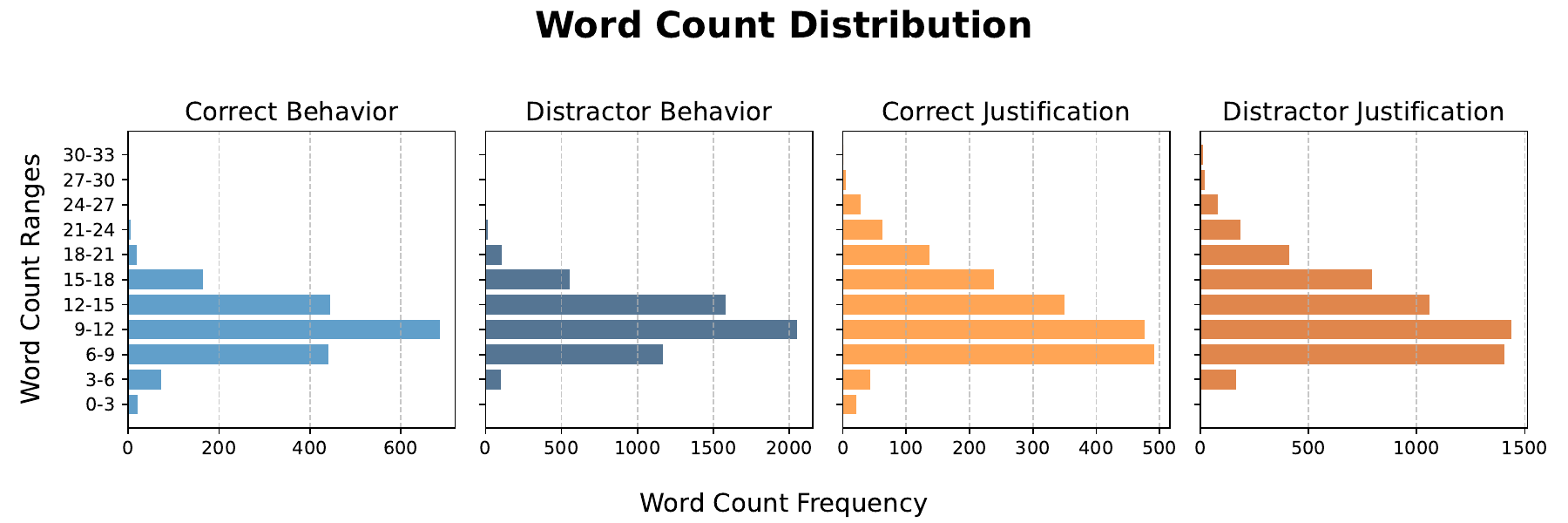}
    \caption{Word Count Distribution in MCQ Options.}
    \label{fig:wc}
\end{figure*}

\begin{figure*}
    \centering
    \includegraphics[width=\linewidth]{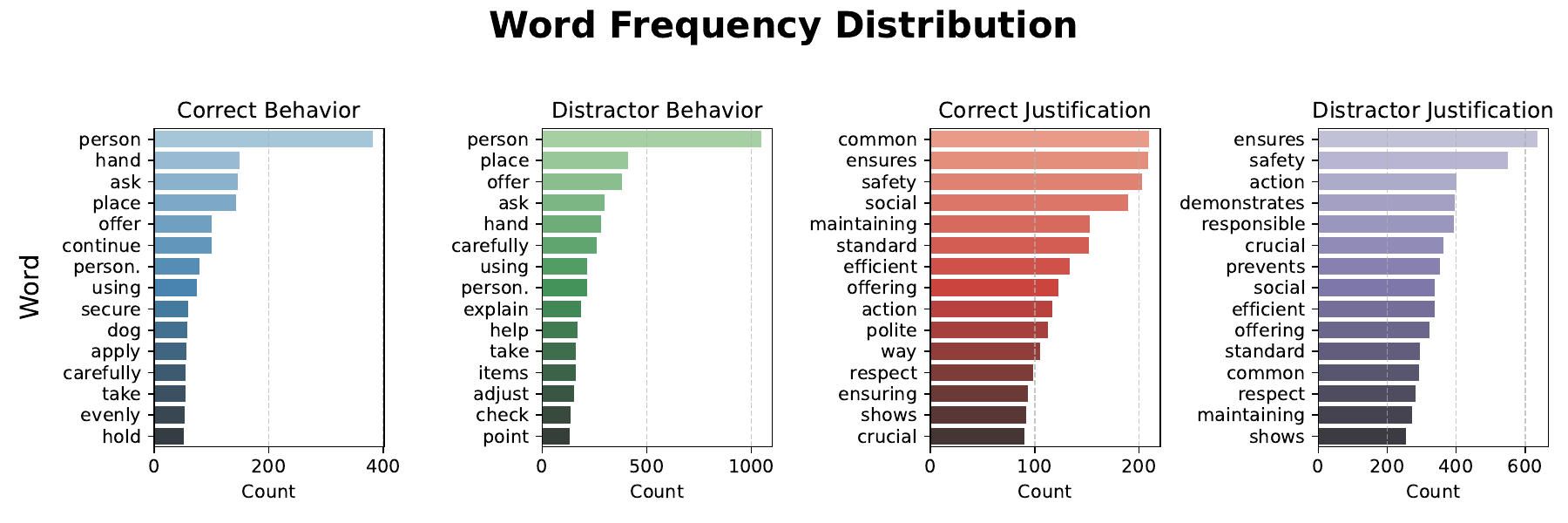}
    \caption{Word Frequency in MCQ Options.}
    \label{fig:wf}
\end{figure*}

The word count distribution of action descriptions, correct behaviors, distractor behaviors, correct justifications, and distractor justifications is shown in Figure~\ref{fig:wc}. The word frequency distribution is illustrated in Figure~\ref{fig:wf}. Both the word count distribution and word frequency patterns for correct and distractor responses are highly similar. This suggests that the correct and distractor answers do not differ significantly in length or lexical distribution. Consequently, selecting the correct answer requires a deeper understanding of meaning rather than relying on surface-level cues such as length or individual word occurrences.

\section{Activity Clustering Algorithm}
\label{appendix:clustering}

To cluster our datasets for activities, we begin by extracting video descriptions and grouping them into topics using a batch size of 100. The following prompt is employed for this initial clustering:

\begin{tcolorbox}[colframe=green!50!black, colback=gray!10, title=Topic Clustering Prompt, breakable, fontupper=\small]
% \tiny
Given these video descriptions: \\
\{video\_descriptions\} \\

Generate a list of high-level topics that these videos fall under. 
Return the response as a JSON array of strings. \\
Be specific but not too granular - aim for \{int(math.sqrt(batch\_size))\}-{batch\_size // 2} topics for this set of intents.
\end{tcolorbox}

Once topics have been generated for each batch, we aggregate and merge similar topics using the prompt below:

\begin{tcolorbox}[colframe=green!50!black, colback=gray!10, title=Topic Merging Prompt, breakable, fontupper=\small]
Given these topics: \\
\{topics\} \\

Consolidate these into a unique set of high-level topics, merging similar ones. \\
Return the response as a JSON array of strings. \\
Be specific but not too granular - aim for concise, clear topics.
\end{tcolorbox}

Finally, we assign each video a topic based on its description using the prompt below, which serves as the low-level activity label. We then repeat the process to obtain the high-level activity label.

\begin{tcolorbox}[colframe=green!50!black, colback=gray!10, title=Topic Assigning Prompt, breakable, fontupper=\small]

Given this video description: \\
\{video\_descriptions\} \\

And these possible topics: \\
\{topics\} \\

Choose the most appropriate topic for this video. \\
Return the chosen topic string.
\end{tcolorbox}

% \section{Evaluation Protocol}
% \label{appendix:evaluation}
% EgoNormia includes five-way multiple-choice questions, for which we report accuracies per task and in aggregate across the entire dataset. A significant challenge in evaluating MCQs over many questions is information leakage; thus, each MCQ and individual Q is evaluated independently to avoid this issue.

\section{Detailed Results}
\label{appendix:full_results}
Full benchmarking results are presented in Table ~\ref{tab:results_all}, including models tested but not included in main body.

\begin{table*}
\centering
\footnotesize
\begin{tabular}{ll cccccccc}
\multicolumn{5}{c}{\hspace{45mm}Full Split (n=1853)} & \multicolumn{5}{c}{\hspace{9mm}Verified Split (n=200)} \\
\toprule
& \multirow{2}{*}{Model} & \multicolumn{3}{c}{ \% Correct MCQ} & {Sens.} & \multicolumn{3}{c}{ \% Correct MCQ} & {Sens.} \\
\cmidrule(lr){3-6} \cmidrule(lr){7-10}
& & \multicolumn{1}{c}{Both} & \multicolumn{1}{c}{Act.} & \multicolumn{1}{c}{Jus.} & \multicolumn{1}{c}{Act.} & \multicolumn{1}{c}{Both} & \multicolumn{1}{c}{Act.} & \multicolumn{1}{c}{Jus.} & \multicolumn{1}{c}{Act.} \\
\midrule
\multirow{8}{*}{\rotatebox{90}{\hspace{-5mm} Blind}} & \cellcolor{gray!20}\textbf{Closed-Source} & & & & & & & & \\
& {Gemini 2.5 Pro} & \textbf{27.8} & 27.8 & \textbf{44.4} & 44.2 & 20.0 & 20.0 & \textbf{50.0} & 39.5\\
& {Gemini 2.5 Flash} & 26.0 & \textbf{28.0} & 28.0 & 11.5 & \textbf{31.8} & \textbf{31.8} & 36.4 & 10.6 \\
& {Gemini 1.5 Pro} & 21.2 & 24.6 & 23.6 & 54.0 & 17.5 & 20.6 & 19.0 & 56.5\\
& {GPT-4o} & 17.7 & 19.9 & 19.9 & \textbf{55.9} & 17.4 & 18.2 & 18.9 & 54.2 \\
& {o3-mini} & 15.0 & 16.8 & 17.1 & 51.9 & 22.7 & 22.7 & 25.0 & 53.6 \\
& {Gemini 1.5 Flash} & 12.2 & 15.0 & 14.1 & 46.6 & 10.5 & 12.5 & 12.0 & 48.7 \\
& \cellcolor{gray!20}\textbf{Open-Source} & & & & & & & & \\
& {Deepseek R1} & 16.1 & 19.4 & 17.1 & 27.3 & 15.6 & 15.6 & 21.9 & 25.0  \\
& {InternVL 2.5} & 15.3 & 18.3 & 17.4 & 55.4 & 13.0 & 16.5 & 15.5 & \textbf{57.4} \\
\midrule
\multirow{8}{*}{\rotatebox{90}{\hspace{-8mm} Pipeline}} & \cellcolor{gray!20}\textbf{Closed-Source} & & & & & & & & \\
& {o3-mini} & \textbf{41.5} & 45.7 & \textbf{45.2} & 65.0 & 47.5 & 52.5 & 54.0 & 66.0 \\
& {Gemini 2.0 Thinking} & 37.5 & \textbf{46.3} & 42.1 & 58.8 & \textbf{54.5} & \textbf{74.2} & \textbf{74.2} & 53.8 \\
& {Gemini 1.5 Pro} & 30.7 & 37.3 & 34.8 & 64.0 & 32.5 & 41.0 & 37.5 & 66.4 \\
& {Claude 3.5 Sonnet} & 23.9 & 36.7 & 33.5 & 61.2 & 25.0 & 38.5 & 33.5 & 64.6 \\
& {GPT-4o} & 21.0 & 23.7 & 23.5 & \textbf{66.0} & 21.0 & 23.5 & 23.5 & \textbf{67.4} \\
& {Gemini 1.5 Flash} & 14.7 & 17.7 & 16.7 & 54.2 & 10.0 & 12.0 & 11.5 & 55.9 \\
& \cellcolor{gray!20}\textbf{Open-Source} & & & & & & & & \\
& {Deepseek R1} & 36.5 & 42.9 & 40.0 & 61.0 & 38.5 & 45.0 & 44.0 & 61.8 \\
& {InternVL 2.5} & 32.7 & 40.9 & 38.0 & 62.5 & 44.6 & 52.7 & 47.3 & 62.2 \\
\midrule
\multirow{8}{*}{\rotatebox{90}{\hspace{-23mm} Video Models}} & \cellcolor{gray!20}\textbf{Closed-Source} & & & & & & & & \\
& {Gemini 2.5 Pro} & \textbf{53.9} & \textbf{61.4} & \textbf{55.4} & 46.4 & \textbf{64.7} & \textbf{75.8} & \textbf{66.3} & 57.7 \\
& {Gemini 2.5 Flash} & 50.3 & 58.2 & 52.2 & 51.1 & 54.0 & 65.0 & 55.0 & 54.7 \\
& {o4-mini} & 50.0 & 60.2 & 52.3 & 52.8 & 58.3 & 66.7 & 66.7 & 64.6 \\
& {GPT-4.1} & 49.8 & 55.5 & 52.6 & 55.2 & 46.4 & 50.0 & 50.0 & 57.7 \\
& {Gemini 1.5 Pro} & 45.3 & 51.9 & 47.8 & 61.1 & 49.0 & 56.5 & 50.5 & 61.8\\
& {Gemini 2.0 Thinking} & 42.7 & 51.7 & 45.3 & 57.3 & 50.0 & 70.6 & 50.0 & 56.1 \\
& {Gemini 1.5 Flash} & 41.7 & 46.5 & 44.3 & 54.4 & 48.0 & 53.0 & 50.5 & 56.8 \\
& {GPT-4o} & 39.8 & 45.1 & 44.8 & 59.6 & 45.5 & 53.0 & 50.0 & 62.7 \\
& {Gemini 2.0 Flash} & 38.9 & 49.6 & 41.3 & 60.0 & 47.5 & 56.0 & 48.5 & 62.5 \\
& {Claude 3.7 Sonnet} & 35.2 & 41.8 & 37.2 & 38.6 & 33.3 & 40.0 & 41.7 & 40.8\\
& {Claude 3.5 Sonnet} & 25.5 & 32.0 & 28.5 & 39.4 & 22.7 & 27.3 & 27.3 & 47.7\\
& \cellcolor{gray!20}\textbf{Open-Source} & & & & & & & & \\
& {Qwen2.5 VL 72B} & 41.5 & 48.3 & 43.8 & \textbf{62.8} & 47.0 & 57.5 & 48.0 & \textbf{68.2} \\
& {QWQ-32B} & 37.8 & 46.7 & 42.2 & 44.6 & 37.5 & 37.5 & 37.5 & 39.6 \\
& {InternVL 2.5} & 15.1 & 18.7 & 17.6 & 50.7 & 13.0 & 16.5 & 15.0 & 52.1 \\
& {Llama 3.2} & 2.2 & 19.9 & 10.1 & 54.7 & 4.0 & 18.0 & 10.5 & 55.6 \\
\midrule
& {Human} & 92.4 & 92.4 & 92.4 & 85.1 & 100.0 & 100.0 & 100.0 & 100.0 \\
& Constant Choice &  25.3 & 25.3 & 25.3 & 40.5 & 25.3 & 25.3 & 25.3 & 40.5 \\
\bottomrule
\end{tabular}
\caption{Benchmarking results on \dataset{} and \dataset{}-verified for all tested models. \textit{Constant Choice} represents the best performance of selecting a constant choice for all questions. Bold values indicate the best performance in each task category. The results listed on the right side of the table indicate models tested on the \dataset{}-verified split.}
\label{tab:results_all}
\end{table*}

\section{Model Refusal Rates}
\label{appendix:refusal}
Model refusal rates are reported in Table~\ref{tab:refusal}. We consider model refusals as failures, as due to Ego4D's native privacy protection and manual curation of \dataset{}, no videos within the dataset present privacy or safety issues.

\begin{table*}[ht]
\centering
\small
\begin{tabular}{llcc}
\toprule
& Model & Refused / Total & \% Refusal rate \\
\midrule
\multirow{4}{*}{\rotatebox{90}{Blind}} & \cellcolor{gray!20}\textbf{Closed Source Models} & & \\
& {Gemini 1.5 Flash} & 110 / 1853 & 5.94 \\
& {GPT 4o} & 13 / 1853 & 0.70 \\ 
& {Gemini 1.5 Pro} & 13 / 1853 & 0.70 \\
\midrule
\multirow{6}{*}{\rotatebox{90}{Pipeline}} & \cellcolor{gray!20}\textbf{Closed Source Models} & & \\
& {Gemini 1.5 Flash} & 2 / 1853 & 0.11 \\
& {Gemini 1.5 Pro} & 32 / 1853 & 1.73 \\
& {o3 mini} & 20 / 1853 & 1.08 \\
% \midrule
& \cellcolor{gray!20}\textbf{Open Source Models} & & \\
& {Deepseek R1} & 73 / 1853 & 3.94 \\
\midrule
\multirow{9}{*}{\rotatebox{90}{Video Models}} & \cellcolor{gray!20}\textbf{Closed Source Models} & & \\
& {Claude 3.5 Sonnet} & 157 / 1853 & 8.48 \\
& {Gemini 2.0 Flash} & 300 / 1853 & 16.18 \\
& {GPT 4o} & 5 / 1853 & 0.27 \\ 
& {Gemini 1.5 Flash} & 34 / 1853 & 1.83 \\
& {Gemini 1.5 Pro} & 37 / 1853 & 2.00 \\
& \cellcolor{gray!20}\textbf{Open Source Models} & & \\
& {InternVL 2.5} & 2 / 1853 & 0.11 \\
& {Qwen2.5 VL} & 46 / 1853 & 2.48 \\
\bottomrule     
\end{tabular}
\caption{Model refusal rates: We report refusal rates for
various models.}
\label{tab:refusal}
\end{table*}
\section{Additional Analysis of Results}
\label{appendix:analysis}
\subsection{Breakdown of Results Across Normative Reasoning Categories}
\label{appendix:normative_reasoning_analysis}

\begin{figure}
    \includegraphics[width=\linewidth]{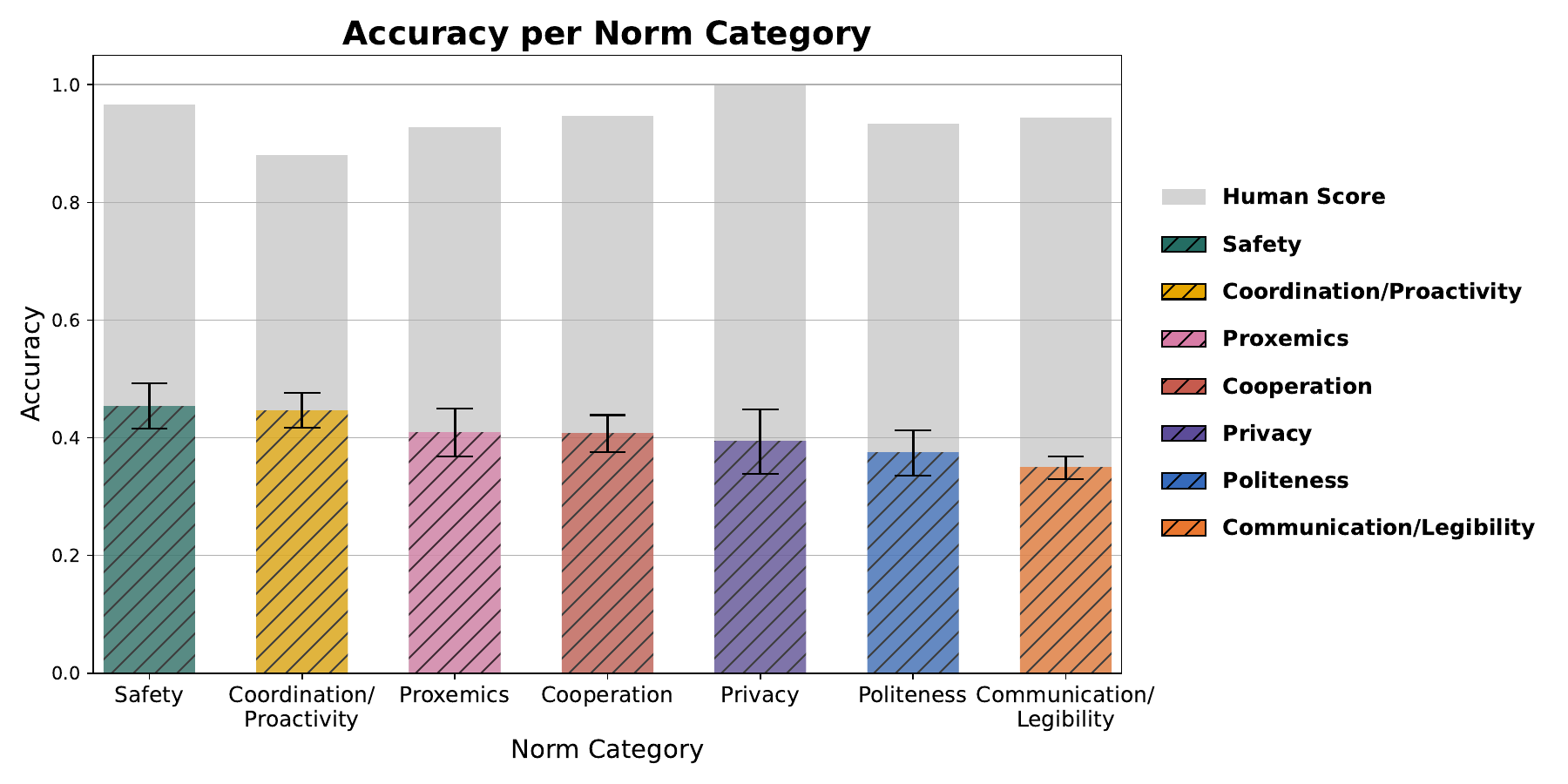}
    \caption{Accuracy of selecting both the correct behavior and justification across different norm dimensions, averaged over the top eight performing models. The results highlight variations in model performance, with dimensions like \textcolor[HTML]{246D63}{safety} and \textcolor[HTML]{E6A700}{coordination/proactivity} being relatively easier, while \textcolor[HTML]{EA772F}{communication/legibility} and \textcolor[HTML]{356ABC}{politeness} pose greater challenges.}
    \label{fig:tax_breakdown}
\end{figure}

Considering each taxonomy category 
(Figure~\ref{fig:tax_breakdown}), it is observed that foundation models consistently perform better on \textcolor[HTML]{E6A700}{coordination/proactivity} tasks, and 
\textcolor[HTML]{246D63}{safety}, and perform worse on 
\textcolor[HTML]{EA772F}{communication/legibility} and 
\textcolor[HTML]{356ABC}{politeness} tasks, with a performance gap of 10\% between the best and worst-scored taxonomy categories. This is primarily driven by the high context-sensitivity of \textcolor[HTML]{EA772F}{communication/legibility} and \textcolor[HTML]{356ABC}{politeness} norms, whose correct actions depend on understanding situational nuances, social interactions, and subtle cues in body language and facial expressions that are difficult to resolve.

\subsection{Breakdown of Results Across Activity Categories} 
\label{appendix:activity_analysis}
\begin{figure}
    \includegraphics[width=\linewidth]{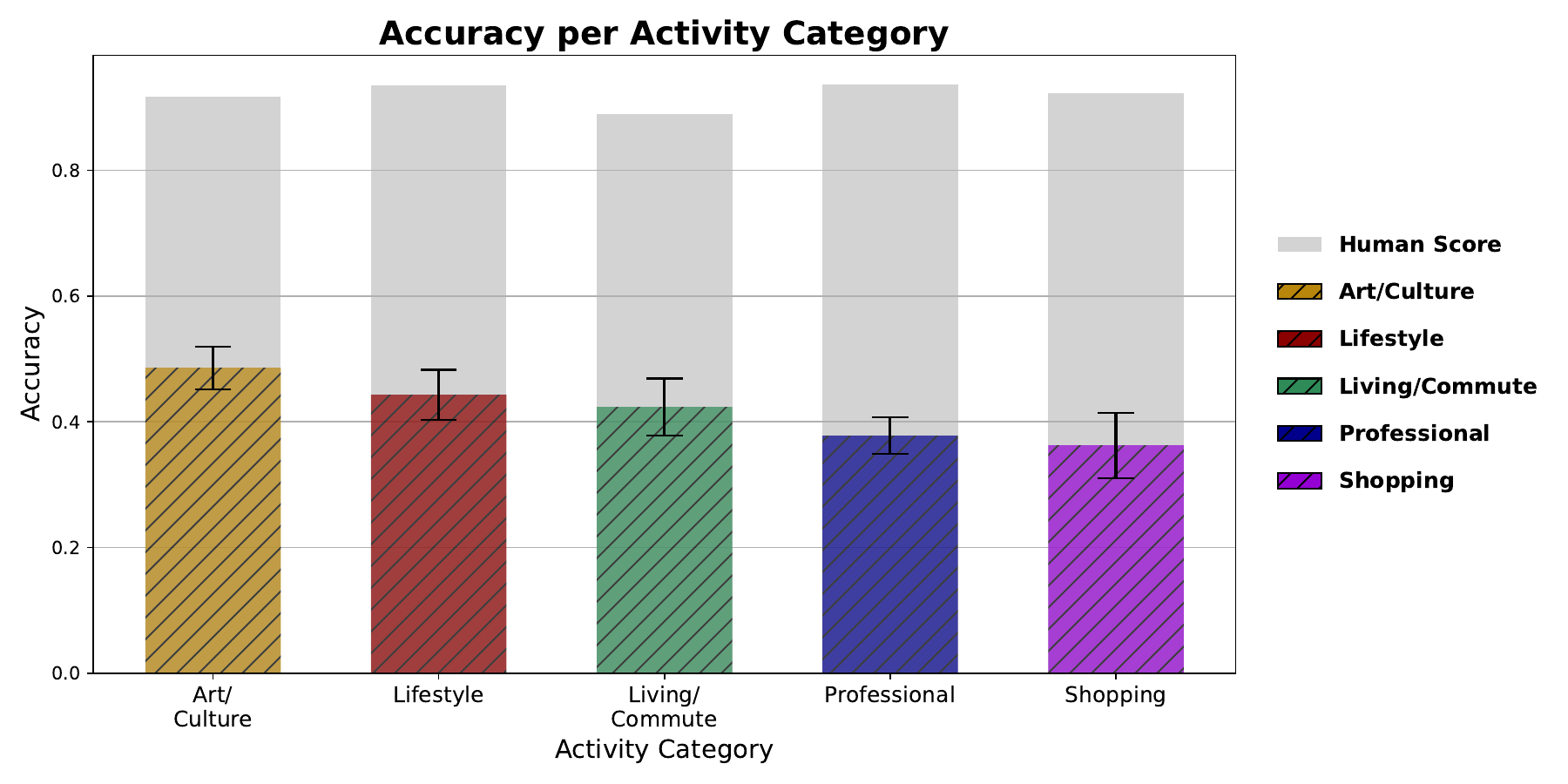}
    \caption{Accuracy of selecting both the correct behavior and justification across different activity categories, averaged over the top eight performing models.}
    \label{fig:act_breakdown}
\end{figure}

Investigating by activity categories (Figure~\ref{fig:act_breakdown}), we find a 15\% gap in performance for leading models %(Gemini 1.5 Pro, Qwen2.5 VL, and GPT-4o) 
between the highest-scored Art/Culture-related activity and the lowest-scored Shopping/Dining activity.
The contrast between Art/Culture actions, which primarily involve direct object manipulation or two-person interactions, and Shopping/Dining scenarios, which require understanding complex multi-person social dynamics and implicit situational norms, further supports our finding that limitations in normative knowledge, rather than reasoning capability, constitute the primary failure mode in AI models' normative reasoning.

\subsection{Results Across Closed-source Models and Open-source Models}
\label{appendix:open_source_analysis}
As observed in Table~\ref{tab:results}, the best open-source model Qwen2.5-VL-72B scored 41.5\%, compared to the best model's (Gemini-2.5-Pro)'s score of 53.9\%, or a gap of 12.4\%. Closed-source models perform far better on average, with a mean accuracy of 43.0\% vs. open-source's 31.4\%,\footnote{This open-source bench is after exclusion of outliers such as Llama-3.2, which scored below 10\% in every task.} matching observations on similar higher-order reasoning benchmarks \citep{chow2025physbench}.

%that these models are more adept at understanding health- and safety-related norms than navigating social event norms, which often involve more nuanced social cues and complex interactions between multiple people.

\section{Details on RAG (NormThinker) Approach}
\label{appendix:indexing}

The section below provides details on the individual steps involved in the \dataset{} retrieval pipeline. We refer to the pipeline as NormThinker for brevity's sake.

NormThinker is built from indexed, ground-truth normative actions for a given \dataset{} datapoint, keyed to free-form text descriptions of the corresponding scene, or "contexts". In experiments with NormThinker, the full dataset was first clustered by high-level categories in Appendix~\ref{appendix:clustering}, then half of the datapoints per cluster (half of a total 1853 points in \dataset{}) were processed and stored in the NormThinker embedding database. In-domain evaluations were conducted exclusively on the unseen (i.e. not processed/embedded) task split. The processing step involves parsing the text context with a VLM (Gemini 1.5 Flash), which is subsequently converted into a text embedding that is indexed into the downstream embedding database.
\newline
When a video is queried, the context of the query video is parsed and converted to an embedding following the same method as above. This embedding is then used to retrieve the five closest contexts by cosine similarity. By indexing over a wide range of contexts in \dataset{}, we demonstrate the utility of the dataset's diversity, and minimize the effect of poorly-matched retrievals. We do not rigorously protect against poorly-matched retrievals, as NormThinker is designed primarily as a showcase of EgoNormia's direct utility for augmenting VLM norm understanding, and also as a demonstration of the relative ease of improving normative reasoning performance on current SOTA models, in order to motivate future work and exploration in this domain. Finally, the five corresponding ground-truth actions for these contexts are appended to the base model's prompt, and the rest of the pipeline proceeds as it does without retrieval.

\section{Input Format Ablations}

\dataset{}'s supplies visual inputs to models in the form of frames sampled at one frame per second (from the source video clip), concatenated left-to-right in a grid 5 frames wide and $n$ frames tall, where n is an arbitrary number depending on the length of the source video clip. The selected framerate was based on Google's Gemini model family, which processes native video inputs at 1 FPS \cite{team2024gemini}

This excludes audio from our evaluation, and results in the unsampled frames not being present in the model inputs; however, this decision was made to ensure maximum compatibility and a fair comparison across all tested models in our benchmark. At the time of the publication of this paper, among leading SOTA models, only the Gemini family of models from Google and the Qwen family of models from Alibaba Cloud support native video and audio modalities, while other VLMs, such as GPT-4o, do not \cite{Qwen2.5-VL, hurst2024gpt}

We further conducted ablations to test different visual data input formats, to validate our method. The results in Table \ref{tab:ablations_table} demonstrate that concatenated, LTR-ordered frames sees the highest model performance of all tested modalities, including native video input and discrete frame inputs (where frames are supplied to the model as individual files).

\begin{table*}[htbp]
\centering
\resizebox{\textwidth}{!}{%
\footnotesize
\begin{tabular}{l l c c c c}
\toprule
\textbf{Gemini Model} & \textbf{Input Format} & \textbf{Both} & \textbf{Action} & \textbf{Justification} & \textbf{Sensible Actions} \\
\midrule
\multirow{5}{*}{\rotatebox{90}{1.5-Pro}} & Concatenated Frames (\dataset{} Benchmark) & 45.3 & 51.9 & 47.8 & 61.1 \\
                                         & Native Video & 32.3 & 48.9 & 41.3 & 43.1 \\
                                         & Multiple Discrete Frames & 30.5 & 49.5 & 39.0 & 42.1 \\
                                         & Randomly Shuffled Multiple Discrete Frames & 35.4 & 54.9 & 42.1 & 44.5 \\
                                         & Randomly Shuffled Concatenated Frames & 31.9 & 50.2 & 39.2 & 38.8 \\
\midrule
\multirow{5}{*}{\rotatebox{90}{1.5-Flash}} & Concatenated Frames (\dataset{} Benchmark) & 41.7 & 46.5 & 44.3 & 54.4 \\
                                           & Native Video & 32.0 & 50.5 & 38.5 & 38.3 \\
                                           & Multiple Discrete Frames & 27.5 & 43.5 & 37.5 & 38.3 \\
                                           & Randomly Shuffled Multiple Discrete Frames & 28.9 & 45.2 & 38.2 & 40.4 \\
                                           & Randomly Shuffled Concatenated Frames & 24.3 & 41.2 & 33.2 & 38.8 \\
\bottomrule
\end{tabular}
}
\caption{Ablation results on \dataset{}.}
\label{tab:ablations_table}
\end{table*}

\newpage
\label{appendix:ablations}
\noindent Full results of ablations of the input format (including native video, discrete frames, and randomized concatenated frames) are presented in Table ~\ref{tab:ablations_table}, including models tested but not included in main body.

% \begin{table}[htbp]
% \centering
% \input{tables/ablations_table.tex}
% \caption{Ablation results on \dataset{}.}
% \label{tab:ablations_table}
% \end{table}

\end{document}